\documentclass[10pt, conference]{IEEEtran}
\IEEEoverridecommandlockouts
\usepackage{cite}
\usepackage{amsmath,amssymb,amsfonts}
\usepackage{algorithmic}
\usepackage{graphicx}
\usepackage{textcomp}
\usepackage{xcolor}
\def\BibTeX{{\rm B\kern-.05em{\sc i\kern-.025em b}\kern-.08em
    T\kern-.1667em\lower.7ex\hbox{E}\kern-.125emX}}
    
\usepackage{bm}
\usepackage[linesnumbered,ruled,vlined]{algorithm2e}
\usepackage{multirow}
\usepackage{subfigure}
\usepackage{bbm}
\usepackage{booktabs}
\usepackage{enumitem}
\usepackage{filecontents}
\usepackage{subfigure} %子图
\usepackage{fancyhdr}
\usepackage{flushend}
\usepackage[misc]{ifsym}

\begin{document}
\title{Saga: Capturing Multi-granularity Semantics from Massive Unlabelled IMU Data}

\author{Yunzhe Li$^*$$^\S$, Facheng Hu$^*$$^\S$\thanks{$^\S$ Yunzhe Li and Facheng Hu contributed equally to this paper.}, Hongzi Zhu$^*$\textsuperscript{\Letter}\thanks{\textsuperscript{\Letter} Hongzi Zhu is the corresponding author of this paper.}, Shifan Zhang$^*$, Liang Zhang$^\dagger$, Shan Chang$^\dagger$, Minyi Guo$^*$ \\ $^*$ \emph{Shanghai Jiao Tong University} \\ $^\dagger$ \emph{Donghua University} \\
\{yunzhe.li, facheng\_hu, hongzi\}@sjtu.edu.cn}

\maketitle

\begin{abstract}
Inertial measurement units (IMUs), have been prevalently used in a wide range of mobile perception applications such as activity recognition and user authentication, where a large amount of labelled data are normally required to train a satisfactory model. However, it is difficult to label micro-activities in massive IMU data due to the hardness of understanding raw IMU data and the lack of ground truth. 
In this paper, we propose a novel fine-grained user perception approach, called \emph{Saga}, which only needs a small amount of labelled IMU data to achieve stunning user perception accuracy. The core idea of Saga is to first pre-train a backbone feature extraction model, utilizing the rich semantic information of different levels embedded in the massive unlabelled IMU data. Meanwhile, for a specific downstream user perception application, Bayesian Optimization is employed to determine the optimal weights for pre-training tasks involving different semantic levels.
We implement Saga on five typical mobile phones and evaluate Saga on three typical tasks on three IMU datasets. Results show that when only using about 100 training samples per class, Saga can achieve over 90\% accuracy of the full-fledged model trained on over ten thousands training samples with no additional system overhead.
\end{abstract}


%%
%% end of the preamble, start of the body of the document source.

\section{Introduction}

Recent years have witnessed a broad range of user perception applications utilizing inertial measurement units (IMUs), including user authentication \cite{shi2021face, xu2020touchpass, zhu2017shakein, jiang2023mauth}, activity recognition \cite{chen2021magx, ouyang2022cosmo, ouyang2021clusterfl}, and health monitoring \cite{cao2021itracku, narayana2021sos}. However, the efficacy of such applications hinges on the availability of expensive and accurately labelled IMU data, which is a requirement often deemed impractical \cite{ouyang2022cosmo, xu2023practically}. Given the huge amount of raw IMU data easily generated on mobile devices, it is natural to ask \emph{whether users of such mobile devices can be well perceived with very few or even no labelled IMU data}, referred to as the IMU-based user perception (IUP) problem. A practical solution to this problem needs to meet the following three rigid requirements. First, the solution can access plenty of unlabelled IMU data but should only require a small amount of labelled data.
Second, the solution should be able to achieve high accuracy over multiple user perception tasks simultaneously to meet the diverse user perception needs. Third, the solution should be lightweight enough to run locally on mobile devices.

%Considering the great success of labelled data pre-training on Natural Language Processing (NLP) area such as BERT \cite{devlin2018bert} and GPT \cite{wu2023brief}, it is natural to ask \emph{whether pre-training on IMU data can also lead to similar imporvements}.

In the literature, pioneer efforts have been made to solve the IUP problem. 
One main direction is to pre-train an IMU feature extraction model via contrastive learning \cite{chen2020simple, qian2022makes, saeed2019multi, haresamudram2021contrastive}, \emph{i.e.}, to obtain IMU representations in certain feature space by maximizing the similarity of the same IMU samples under different transformations. However, it is difficult to find good transformations for IMU data \cite{qian2022makes, xu2023practically,li2025prism}. Another direction is to pre-train an IMU feature extraction model using generative methods (\emph{e.g.}, BERT \cite{devlin2018bert} and GPT \cite{wu2023brief}), where models are trained by predicting those masked data samples or future data points \cite{xu2021limu, cui2024harfmr, miao2024spatial, haresamudram2020masked}. However, all existing methods using generative pre-training methods simply treat one single IMU data sample point as a word in a text, ignoring the inherent differences between text and IMU data, unable to achieve a satisfactory performance on the IUP problem. As a result, to the best of our knowledge, there is no successful solution to the IUP problem.

\begin{figure}
	\includegraphics[width=\linewidth]{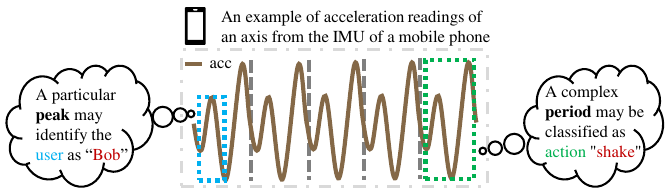}
	\caption{Illustration of the semantics of different levels embedded in IMU data, where particular shapes and complex periods in IMU data contain rich information such as identification or/and activity type of a user.}
	\label{fig:motivation}
    %\vspace{-0.5cm}
\end{figure}

In this paper, we propose an effective IMU-based fine-grained user perception approach, called \emph{Saga}, which makes full use of massive unlabelled IMU data and can achieve stunning user perception accuracy at a low data-labelling cost. As illustrated in Figure \ref{fig:motivation}, we have the key observation that particular shapes and complex periods in the IMU data contain rich information such as identification or/and activity type of a user. If a pre-trained model can successfully predict IMU shapes of different scales and periods for different users, such a model has learned supreme representations of IMU data for various downstream user perception tasks. Therefore, the core idea of Saga is to first pre-train a backbone feature extraction model by predicting masked IMU data segments of different levels, enforcing the backbone to characterize multi-granularity semantics embedded in the massive unlabelled IMU data. Meanwhile, for a specific downstream user perception application, an iterative Bayesian Optimization process is employed in finding the optimal weights among these pre-training tasks, using a small amount of labelled IMU data.

%
%For example, the semantics embedded in periods of IMU data are significant for the activity recognition task, as they encapsulate information pertinent to the activities themselves. Conversely, for a user authentication task, these semantics behind periods become noise, while the minutiae within a given interval, such as the shape of peak, emerge as crucial distinguishing factors for identifying a specific user. Conversely, for a user authentication task, these period-level semantics (\emph{e.g.}, activities within a period) become noise. The sub-period-level semantics (\emph{e.g.}, the shape of peak) emerge as crucial distinguishing factors for identifying a specific user.
%Based on the above observation, we choose to pre-train Saga on unlabelled IMU data to learn multi-level semantics within the IMU data to satisfy specific focuses of diverse downstream tasks.

The design of Saga faces two challenges. First, unlike the semantics embedded in text or images, which are easy for humans to explain, the semantics embedded in IMU data are hard for human comprehension. Therefore, it is challenging to design appropriate pre-training tasks that can well reflect the comprehensive IMU data semantics. To address this challenge, we delve into analyzing the semantics of IMU data and observe that distinct actions of different people show particular shapes of different periods. In line with this observation, we consider four pre-training tasks of different levels, \emph{i.e.}, sensor level, point level, sub-period level, and period level. In each pre-training task, certain IMU data points are masked and the backbone model is asked to regress the values of these masked points. Specifically, in a sensor-level pre-training task, we randomly mask all data points on one axis of the IMU; in a point-level pre-training task, we randomly mask a time window of data points on all axes of the IMU; in a sub-period-level pre-training task, we randomly mask all data points between a pair of key points (\emph{e.g.}, peak and valley points, or crossing-zero points); in a period-level pre-training task, we randomly mask a main period identified in the IMU data. As a result, multi-level semantics embedded in the IMU data can be extracted and learned by the pre-trained backbone model.

Second, given a specific downstream user perception application, it is hard to determine the optimal weights among these pre-training tasks so that the derived backbone performs best. Indeed, the weight designation among pre-training tasks is a complex decision-making process, which can be proved to be NP-hard \cite{bartunov2020continuous}. To solve this problem at a low cost, we design a weight search strategy based on Bayesian Optimization \cite{shahriari2015taking}. Specifically, a performance model based on Gaussian Process \cite{schulz2018tutorial} is employed to capture the influence of weights assigned to various pre-training tasks on downstream tasks training. During each training iteration, the optimal weights in the view of the performance model are first selected to train the backbone. Then, a small number of labelled training samples are used to end-to-end fine-tune and verify the backbone and a pre-trained downstream classifier. The current and all previous validation outcomes are collectively utilized to further refine the performance model. This iterative process repeats until the validation outcomes converge or a pre-defined training budget is reached. In this way, a set of satisfactory weights of pre-training tasks can be found at a low cost.

We implement Saga on five mobile phones (\emph{i.e.}, Mi 6, Pixel 3 XL, Honor v9, Mi 10 and Mi 11) and evaluate the performance of Saga on three public IMU datasets (\emph{i.e.}, HHAR \cite{stisen2015smart}, Motion \cite{malekzadeh2019mobile}, and Shoaib \cite{shoaib2014fusion}). A total of three types of IMU perception tasks, including activity recognition (AR), user authentication (UA), and device placement recognition (DP) are considered. Results show that Saga can perform extremely well especially when only a small number of training samples are available, with an increase in terms of perception accuracy up to 51.6\%. 
On average, Sage can outperform the state-of-the-art methods in terms of relative accuracy by 11.8\% when only 80 training samples are provided. When Saga only uses 100 labelled training samples per class, it can achieve an accuracy of over 90\% relative to the IUP accuracy using all labelled data. Moreover, Saga is lightweight and can be easily deployed on mobile devices.

We conclude the contributions of Saga as follows: 
\begin{itemize}
    \item A novel backbone pre-training scheme for IMU data is proposed, consisting of a set of four pre-training tasks particularly designed for capturing semantics of different granularities;
    \item An effective weight searching strategy based on Bayesian Optimization is introduced to search satisfactory weights of pre-training tasks at a low cost;
    \item Saga is implemented on various devices and extensive experiments over multiple IMU datasets are conducted, results of which indicate the efficacy of Saga.
\end{itemize}

\section{Related Work}

\subsection{Machine Learning for IMU data}

Machine learning has been widely used for sensing applications based on IMU data \cite{shi2021face, xu2020touchpass, zhu2017shakein, zhang2020smartso}. Traditional machine learning models are first used for IMU data, \emph{e.g.}, Hidden Markov Model (HMM) \cite{xu2016air}, Support Vector Machine (SVM) \cite{zhu2017shakein, zhang2020smartso} and Dynamic Time Warping (DTW) \cite{xu2021novel}. These methods are easily deployed and work well with a small number of data samples. However, they all rely on manually designed features. For automatic feature extraction, deep-learning-based models are used to design models for IMU data. CNN-based models \cite{yang2015deep, ding2019deep} with a strong feature extraction capability are first used for inference on IMU data. RNN-based models \cite{li2021cross, xu2021limu} are also useful with a strong generalization capability. Transformer-based models can also be used for sensing with IMU data  \cite{xu2021limu, qian2022makes}.

\subsection{Pre-training on Unlabelled Data}

\begin{figure*}
    \centering
	\includegraphics[width=\linewidth]{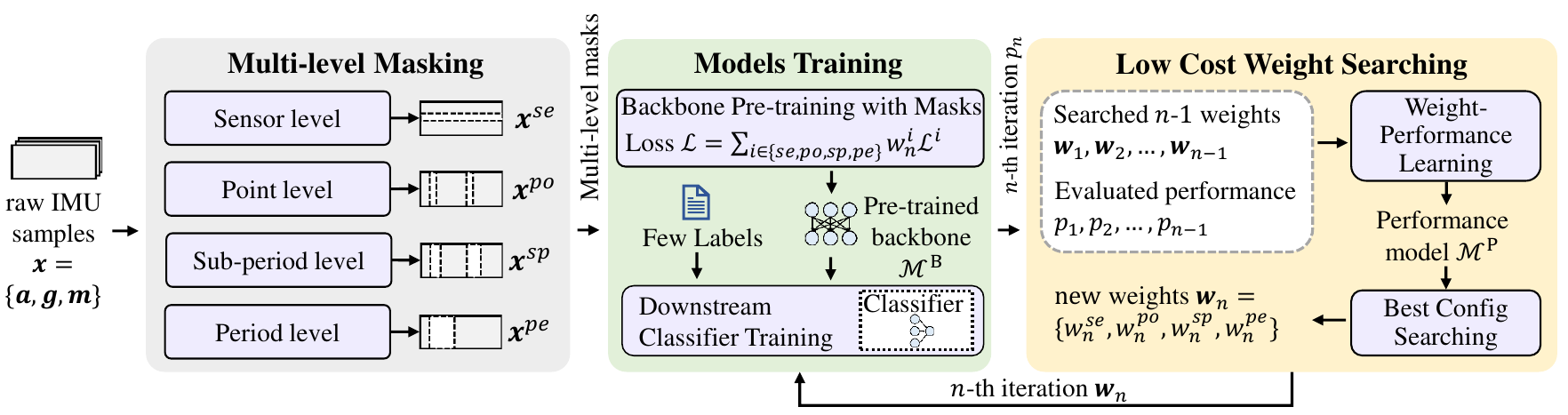}
   % \vspace{-0.3cm}
	\caption{Overview of Saga, where original IMU samples are masked at four levels and pre-trained with weighted loss. The weights of the losses are optimized by Bayesian Optimization.}
   % \vspace{-0.5cm}
	\label{fig:overview}
\end{figure*}

Pre-training on unlabelled data has been a popular topic in the deep learning community for the great success of pre-trained large language models (LLMs) such as ChatGPT \cite{wu2023brief}. In general, model pre-training aims to learn common representations among tasks from as much data as possible to reduce the difficulty of downstream task training. 
One kind of pre-training method is supervised pre-training, which utilizes a large amount of labelled data for one specific task for model pre-training \cite{zoph2020rethinking, he2019rethinking, zhou2022mogde}. For example, pre-training the backbones of object detection and segmentation models on ImageNet classification \cite{deng2009imagenet} was once a common practice \cite{he2019rethinking}. However, supervised pre-training is hard to scale up because of the lack of enough annotated data in many fields. In contrast, unsupervised pre-training tries to annotate the unlabelled samples automatically by designing specific pre-training tasks. One mainstream unsupervised pre-training task is contrastive task \cite{chen2020simple, caron2018deep}. The contrastive tasks aim to partition the data samples into several classes by clustering \cite{caron2018deep} or data augmentation \cite{chen2020simple, chen2021exploring, grill2020bootstrap, he2020momentum}, and then the partition results can be utilized for supervision. Another mainstream unsupervised pre-training task is generative task \cite{wu2023brief, devlin2018bert}, which aims to use part structure in a sample for supervision.

\subsection{Unlabelled Pre-training for IMU data}

There are also several research on IMU data pre-training \cite{saeed2019multi, haresamudram2021contrastive, xu2021limu, qian2022makes, haresamudram2024large}. TPN \cite{saeed2019multi} first tries to pre-train on IMU data by introducing multiple transformations on IMU data to use these transformations as supervision. CL-HAR \cite{qian2022makes} further explored the effectiveness of different contrastive learning methods on IMU data. However, it is difficult to choose a good transformation for IMU data pre-training \cite{qian2022makes, xu2023practically}. As a result, these contrastive-learning-based IMU pre-training methods cannot ensure stable performance on various datasets. LIMU-BERT \cite{xu2021limu} introduces BERT structure and Masked Language Model (MLM) task originally designed on text \cite{devlin2018bert} for pre-training on IMU data, which is SOTA IMU pre-training method among MLM-based methods \cite{cui2024harfmr, miao2024spatial, haresamudram2020masked}. However, existing methods simply view each IMU sample point as a word in the text, ignoring semantics in IMU data. This makes it hard to learn targeted feature representations for various downstream tasks. In contrast, the proposed Saga method tries to delve into IMU data by semantics with diverse granularity and adapt the weights of different pre-training tasks for every downstream task, which is efficient and adaptable for various downstream tasks.

\section{Overview of Saga}

The core idea of Saga is to partition the IMU data according to the semantics of IMU data and then the partitioned semantics are masked as supervised information of Deep Neural Networks (DNNs) for unlabelled pre-training. The supervisions of semantics on different levels are weighed as the final supervision. The weights of different semantics for different downstream tasks are searched efficiently based on Bayesian Optimization \cite{shahriari2015taking}. To this end, as illustrated in Figure \ref{fig:overview}, Saga consists of the following three parts.

\textbf{Multi-level Masking (MM).} Given a set of unlabelled IMU samples $\bm{x} \in \mathbb{R}^{L^{win}*3N^{se}}$, where $\mathbb{R}$ denotes real number, $L^{win}$ denotes the length of the slicing window and $N^{se}$ denotes the number of sensors in IMU, MM aims to generate masks based on IMU semantics. Specifically, MM masks $\bm{x}$ based on four levels of IMU semantics, \emph{i.e.}, sensor-level mask $\bm{x}^{se}$, point-level mask $\bm{x}^{po}$, sub-period-level mask $\bm{x}^{sp}$ and period-level mask $\bm{x}^{pe}$, for extracting semantics of different levels.

\textbf{Models Training (MT).} Given masks of different levels for multi-granularities, \emph{i.e.}, $\bm{x}^{se}$, $\bm{x}^{po}$, $\bm{x}^{sp}$ and $\bm{x}^{pe}$, MT first pre-trains a backbone, denoted as $\mathcal{M}^B$, with the pre-training tasks of reconstructing masked IMU samples to the original $\bm{x}$. The losses of different pre-training tasks are individually computed and weighed with a weight $\bm{w}=\{w^{se}, w^{po}, w^{sp}, w^{pe}\}$, where $w^{se}$, $w^{po}$, $w^{sp}$, $w^{pe}$ denote the weight for $\bm{x}^{se}$, $\bm{x}^{po}$, $\bm{x}^{sp}$ and $\bm{x}^{pe}$, respectively. $\bm{w}$ will be searched efficiently further.
Then, $\mathcal{M}^B$ is further trained with corresponding classifiers for downstream tasks on a few labelled samples.

\textbf{Low Cost Weight Searching (LWS).} Given all possible weights, denoted as $\mathbb{W} \in \mathbb{R}^{N^{w}}$, where $N^{w}$ denotes the number of weights, historical searched pre-training weights, denoted as $\{\bm{w}_1, \bm{w}_2, \cdots, \bm{w}_{n-1}\}$, and corresponding performance, denoted as $\{p_1, p_2, \cdots, p_{n-1}\}$, LWS aims to search the best config for the considered downstream task with a limited searching budget. To this end, LWS utilizes a performance model based on Gaussian Process (GP) \cite{schulz2018tutorial}, denoted as $\mathcal{M}^{\text{P}}$, to learn the relationship between weight $\{\bm{w}_1, \bm{w}_2, \cdots, \bm{w}_{n-1}\}$ and performance $\{p_1, p_2, \cdots, p_{n-1}\}$. During each iteration, LWS selects the weight that the current performance model $\mathcal{M}^{\text{P}}$ considers to be the best, denoted as $\bm{w}_n$. The performance $p_n$ of weight $\bm{w}_n$ will be further utilized to refine performance model $\mathcal{M}^{\text{P}}$. This process will iterate until the search budget is exhausted or the validation results converge.

\section{Multi-level Masking}

% To effectively utilize the semantics embedded in IMU data for generative pre-training, these semantics should first be masked in IMU data. In Saga, four levels of semantics are masked.

\subsection{IMU Data Preprocessing}

Before masking the raw IMU samples, we first preprocess these IMU samples to find the key points (\emph{i.e.}, peak and valley points) and the main period, as illustrated in Figure \ref{fig:semantics}. The key points can partition the IMU data into several sub-periods, and the main periods contained in IMU data are related to the corresponding actions when collecting the IMU data. 

\subsubsection{Finding Key Pionts in IMU Data}

\begin{figure*}
\centering
  \begin{minipage}[t]{0.3\linewidth}
    \centering
    \includegraphics[width=\linewidth]{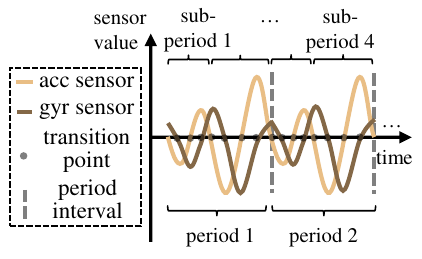}
    %\vspace{-0.8cm}
	\caption{Illustration of semantics in IMU data: 1) The IMU data has periodicity; 2) The IMU data's three axes are time-dependent, \emph{i.e.}, experiencing key points simultaneously.}
    %\vspace{-0.5cm}
	\label{fig:semantics}
  \end{minipage}
  \hspace{0.3cm}
  \begin{minipage}[t]{0.3\linewidth}
    \centering
    \includegraphics[width=0.8\linewidth]{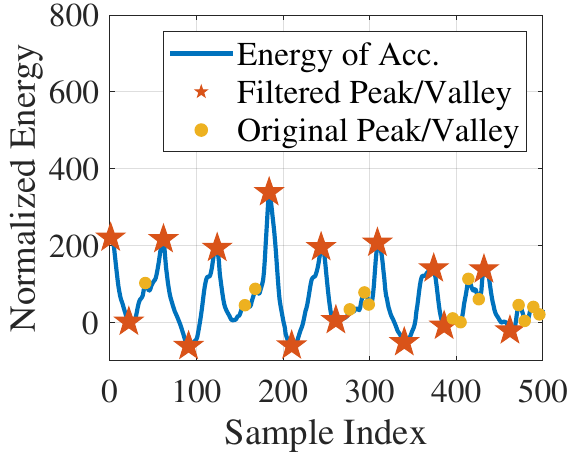}
    %\vspace{-0.4cm}
    \caption{Illustration of finding key points, where the peaks and valleys are affected by small spikes, and the designed filtering method can filter the "fake" points.}
   % \vspace{-0.5cm}
    \label{fig:peak}
  \end{minipage}
  \hspace{0.3cm}
  \begin{minipage}[t]{0.3\linewidth}
    \centering
    \includegraphics[width=0.8\linewidth]{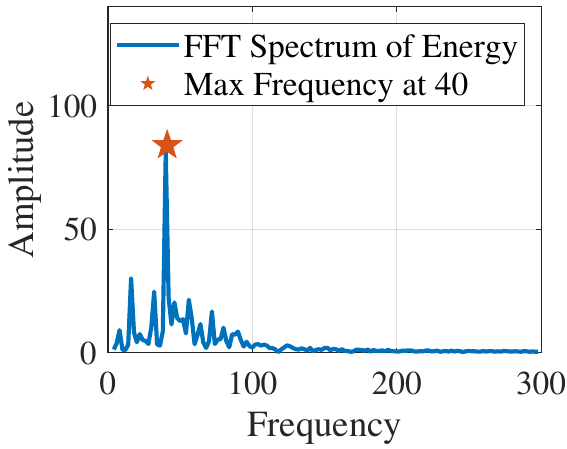}
   % \vspace{-0.4cm}
    \caption{Illustration of key points for finding periods, where the period associated with the maximum amplitude is used as the period of the whole IMU sample.}
    % \vspace{-0.5cm}
    \label{fig:fft}
  \end{minipage}
\end{figure*}

In general, the peaks and valleys in IMU data can be defined as local maximum points and local minimum points in IMU data, respectively. However, as illustrated in Figure \ref{fig:peak}, the collected IMU data contains small spikes, causing each sub-period to be very fragmented. In order to extract the semantics of complete sub-periods, we choose to discard too small peaks and valleys or two peaks and valleys that are too close. Specifically, given an IMU data $\bm{x} = \{x_{1}, x_{2}, \cdots, x_{L^{win}}\}$, we first compute the energy of $\bm{x}$, denoted as $\bm{e} = \{e_1, e_2, \cdots, e_{L^{win}}\}$, to extract the actual motions, defined as: $e_{i} = a_{i1}^2 + a_{i2}^2 + a_{i3}^2$, where $e_{i}$ denotes the energy in $i$-th point, and $a_{i1}$, $a_{i2}$ and $a_{i3}$ denotes the acceleration in all three axes, respectively. Note that the IMU data's three axes are time-dependent, meaning a crossing zero point in acceleration recordings may correspond to a peak in gyroscope recordings. Therefore, such a transformation will not confuse key points in raw IMU data.
Then, each local maximum or local minimum in $\bm{e}$ is identified: $\bm{e}_{p^c} = \{i | e_i \ge e_{i-1} \text{ and } e_i \ge e_{i+1}\}$ and $\bm{e}_{v^c} = \{i | e_i \le e_{i-1} \text{ and } e_i \le e_{i+1}\}$, where $\bm{e}_{p^c}$ and $\bm{e}_{v^c}$ denote the set of local maximum and local minimum in $\bm{e}$, respectively. The points in $\bm{e}_{p^c}$ and $\bm{e}_{v^c}$ are filtered by the following two conditions:
\begin{equation}
	e_i \circ e_j \text{ for all } j \text{ such that } |i-j| \le w ,
\end{equation}
\begin{equation}
	|i_c - j_c| \ge d \text{ for any } i_c, j_c,
\end{equation}
where $\circ$ denotes the relation symbol, \emph{i.e.}, $\ge$ for $\bm{e}_{p^c}$ and $\le$ for $\bm{e}_{v^c}$; $i_c$ and $j_c$ denote index of $\bm{e}_{p^c}$ or $\bm{e}_{v^c}$. The filtered peaks and valleys, denoted as $\bm{e}_{p}$ and $\bm{e}_{v}$, are finally defined as the identified key points for defining the sub-periods.

\subsubsection{Finding the Main Period in IMU Data}

To find the main period in IMU data, we first employ Fourier transform \cite{duhamel1990fast} to identify the periodicity within the IMU data and subsequently partition and mask the IMU data based on this identified period, as illustrated in Figure \ref{fig:fft}. Specifically, given a sequence of the energy of $\bm{x}$, \emph{i.e.}, $\bm{e} = \{e_1, e_2, \cdots, e_{L^{win}}\}$, we initially perform a Fourier transform on $\bm{e}$, transforming it from the time domain to the frequency domain, \emph{i.e.}, 
$E(f) = \int_{-\infty}^{\infty} e(t) \cdot e^{-j2\pi ft} dt$, 
% \begin{equation}
% 	E(f) = \int_{-\infty}^{\infty} e(t) \cdot e^{-j2\pi ft} dt,
% \end{equation}
where $e(t)$ denotes $\bm{e}$ in time domain and $E(f)$ denotes the energy of $\bm{e}$ in different frequencies.
Subsequently, we identify the period associated with the maximum amplitude in the frequency domain, denoted as 
$f_{\text{max}} = \arg \max |E(f)|$.
% \begin{equation}
	% f_{\text{max}} = \arg \max |E(f)|.
% \end{equation} 
The corresponding period of $f_{\text{max}}$, denoted as $T_{\text{main}}$, is then defined as the main period of the IMU data: $T_{\text{main}} = {1}/{f_{\text{max}}}$.

\subsection{Sensor-level Masking}

Sensor-level mask $\bm{x}^{se}$ is designed in order to extract the semantics related to the underlying device, where the recordings of one specific axis of sensors will be masked. Specifically, several masking indices, denoted as $m^{se}$ are first randomly sampled from a uniform distribution, denoted as $U[0,3N^{se})$. Then, denote values in an IMU data point as $x_{i} = \{x_{i}^1, x_{i}^2, \cdots, x_{i}^{3N^{se}}\}$ in $\bm{x}$, where $x_{i}^q$ denotes the value of the $q$-th value in an IMU data point $x_{i}$. For each $x_{i}$, the value at indices in $m^{se}$ is masked as: 
\begin{equation}
    x_i^q = x_i^q \cdot (1 - \mathbbm{1}_{m^{se}}(q)),
\end{equation}
where $\mathbbm{1}_{m^{se}}(q)$ is the indicator function for the set $m^{se}$. This function is defined such that $\mathbbm{1}_{m^{se}}(q) = 1$ when $q \in m^{se}$ and $\mathbbm{1}_{m^{se}}(q) = 0$ when $q \notin m^{se}$.
% \begin{equation}
% 	x_{i}^q = \begin{cases}
% 		0, & \text{if $q \in m^{se}$;}\\
% 		x_{i}^q, & \text{if $q \notin m^{se}$.}
% 	\end{cases}
% \end{equation}

\subsection{Point-level Masking}

% \begin{figure}[]
% 	\centering
% 	\includegraphics[width=0.5\linewidth]{figures/semantics.pdf}
% 	\caption{Illustration of semantics in IMU data: 1) (Periodicity) The IMU data collected by mobile devices has periodicity, which comes from the periodicity of the mobile device's own movement; 2) (Axis correlation) Although IMU data has many axes, the data on these axes is time-dependent, meaning they all experience turning points simultaneously.}
% 	\label{fig:semantics}
% \end{figure} 

Piont-level mask $\bm{x}^{po}$ aims to explore the basic structure in IMU data at the level of data points. An intuitive way for point point-level masking is to randomly select several points discretely and mask them as zero. However, IMU data is continuous in time (shown in Figure \ref{fig:semantics}). As a result, masks of discrete IMU data points can be easily reconstructed by interpolating between surrounding IMU data points \cite{xu2021limu}. As a result, we choose to use Span Masking \cite{joshi2020spanbert, xu2021limu} to mask continuous IMU data points. Specifically, given a unmasked IMU data $\bm{x} = \{x_{1}, x_{2}, \cdots, x_{L^{win}}\}$, a masking length $l^{po}$ is sampled from a geometric distribution $Geo(p)$ clipped at the maximum length, denoted as $l_{max}$: $P(c=k) = (1-p)^{k-1}p, \text{ for } c \in [1, l_{max}],$
% \begin{equation}
% 	P(c=k) = (1-p)^{k-1}p, \text{ for } c \in [1, l_{max}],
% \end{equation}
where $P(c=k)$ denotes the probability of length $k$ being selected in a geometric distribution $Geo(p)$; $p$ denotes the probability of success in $Geo(p)$. Then, a masking starting point $s$ is randomly sampled from a uniform distribution from $0$ to $L^{win}$, \emph{i.e.}, $U[0,L)$. The masked indices are therefore to be $m^{po} = [s, s+l^{po})$. Finally, the data point $x_i$ with indices $m^{po}$ is masked as: 
\begin{equation}
    x_{i} = x_{i} \cdot (1 - \mathbbm{1}_{[s, s + l^{po})}(i)),
\end{equation}
where $\mathbbm{1}_{[s, s + l^{po})}(i)$ is the indicator function that equals $1$ when $i$ is within the interval $[s, s + l^{po})$ and equals $0$ when $i$ is outside the interval.
% \begin{equation}
% 	x_{i} = \begin{cases}
% 		0, & \text{if $i \in [s, s+l^{po})$;}\\
% 		x_{i}, & \text{if $i \notin [s, s+l^{po})$.}
% 	\end{cases}
% \end{equation}

\subsection{Sub-period-level Masking}

Sub-period-level mask $\bm{x}^{sp}$ aims to extract the semantics of action compositions in IMU data. Given the filtered peaks $\bm{e}_{p}$ and valleys $\bm{e}_{v}$, the IMU data $\{x_{1}, x_{2}, \cdots, x_{L^{win}}\}$ can be partitioned into several sub-periods, denoted as $\{x^{sp}_1, x^{sp}_2, \cdots, x^{sp}_{N^{sp}}\}$, where $N^{sp}$ denotes the number of the partitioned sub-periods. A masking index $i^{sp}_m$ is randomly sampled from a uniform distribution from 0 to $N^{sp}$, \emph{i.e.}, $U[0,N^{sp})$. The masked indices are therefore to be the indices of the sampled $x_{i^{sp}_m}^{sp}$, denoted as $m^{sp}$. Finally, the data point with indices $m^{sp}$ is masked as:
\begin{equation}
    x_{i} = x_{i} \cdot (1 - \mathbbm{1}_{m^{sp}}(i)),
\end{equation}
where $\mathbbm{1}_{m^{sp}}(i)$ is the indicator function that $\mathbbm{1}_{m^{sp}}(i) = 1$ when $i \in m^{sp}$ and $\mathbbm{1}_{m^{sp}}(i) = 0$ when $i \notin m^{sp}$.

% \begin{equation}
% 	x_{i} = \begin{cases}
% 		0, & \text{if $i \in m^{sp}$;}\\
% 		x_{i}, & \text{if $i \notin m^{sp}$.}
% 	\end{cases}
% \end{equation}

\subsection{Period-level Masking}

Period-level mask $\bm{x}^{pe}$ aims to extract the semantics of the whole periodic actions in IMU data. 
Given the identified main period in IMU data, the IMU data can then be partitioned into multiple periods based on this identified period, denoted as $\{\bm{x}_1^{pe}, \bm{x}_2^{pe}, \cdots, \bm{x}_{N^{pe}}^{pe}\}$, where $N^{pe}$ denotes the number of periods: $\bm{x}_i^{pe} = \{x_j| j \ge \max \{0, (i-1) \cdot T_{\text{main}}\} \text{ and } j< i \cdot T_{\text{main}}\}.$
% \begin{equation}
% 	\bm{e}_i^{pe} = \{e_j| j \ge \max \{0, (i-2) \cdot T_{\text{pri}}\} \text{ and } j<(i-1) \cdot T_{\text{pri}}\}.
% \end{equation}
Finally, an index $m^{pe}$ within $N^{pe}$ is randomly sampled from a uniform distribution, denoted as $U[0, N^{pe})$. The data points within $m^{pe}$-th sub-sequence, the set of whose indices denoted as $m^{pe}$, is masked:
\begin{equation}
    x_{i} = x_{i} \cdot (1 - \mathbbm{1}_{m^{pe}}(i)),
\end{equation}
where $\mathbbm{1}_{m^{pe}}(i)$ is the indicator function that equals $1$ when $i \in m^{pe}$ and $\mathbbm{1}_{m^{pe}}(i) = 0$ when $i \notin m^{pe}$.

% \begin{equation}
% 	x_{i} = \begin{cases}
% 		0, & \text{if $i \in m^{pe}$;}\\
% 		x_{i}, & \text{if $i \notin m^{pe}$.}
% 	\end{cases}
% \end{equation}

\section{Models Training}

\subsection{Backbone Pre-training with Masks}

With the four-level masks from four-level IMU semantics, we pre-train the backbone model $\mathcal{M}^{\text{B}}$ by reconstructing the masked IMU samples, \emph{i.e.}, predicting the masks $\bm{x}^{se}$, $\bm{x}^{po}$, $\bm{x}^{sp}$ and $\bm{x}^{pe}$. Specifically, for any $\bm{x}^*$ in $\bm{x}^{se}$, $\bm{x}^{po}$, $\bm{x}^{sp}$ and $\bm{x}^{pe}$, $\mathcal{M}^{\text{B}}$ is allowed to reconstruct $\bm{x}^*$ to $\bm{x}$. The Mean Square Error (MSE) loss is utilized for pre-training: $\mathcal{L}^* = \frac{1}{N} \sum_{i}^{L^{win}}(x_i - \mathcal{M}^{\text{B}}(x_i^*))^2$, 
% \begin{equation}
% 	\mathcal{L}^* = \frac{1}{N} \sum_{i}^{L^{win}}(x_i - \mathcal{M}^0(x_i^*))^2,
% \end{equation}
where $x_i^*$ denotes the $i$-th sample in $\bm{x}^*$; 	$\mathcal{L}^*_{\text{mse}}$ denotes the loss for 4 level of masks, \emph{i.e.}, $\mathcal{L}^{se}_{\text{mse}}$, $\mathcal{L}^{po}_{\text{mse}}$, $\mathcal{L}^{sp}_{\text{mse}}$ and $\mathcal{L}^{pe}_{\text{mse}}$. The four losses are combined in a weighted average manner:
\begin{equation}
	\mathcal{L} = w^{se} \cdot  \mathcal{L}^{se}_{\text{mse}} + w^{po} \cdot \mathcal{L}^{po}_{\text{mse}} + w^{sp} \cdot \mathcal{L}^{sp}_{\text{mse}} + w^{pe} \cdot \mathcal{L}^{pe}_{\text{mse}},
\end{equation}
where $\mathcal{L}$ denotes the total loss to be optimized for pre-training the backbone model $\mathcal{M}^{\text{B}}$.

\subsection{Downstream Classifier Training}

After pre-training, we then fine-tune the pre-trained model $\mathcal{M}^{\text{B}}$ on the specific downstream task. Specifically, denoting the classifier model for the downstream task as $\mathcal{M}^{\text{C}}$, the available few labelled IMU samples on the downstream task as $\bm{x}^{\text{C}} = \{x^{\text{C}}_1, x^{\text{C}}_2, \cdots, x^{\text{C}}_N\}$ and $\bm{y}^{\text{C}} = \{y^{\text{C}}_1, y^{\text{C}}_2, \cdots, y^{\text{C}}_N\}$, respectively, $\mathcal{M}^{\text{C}}$ is trained with cross-entropy loss:
\begin{equation}
	\mathcal{L}^{\text{C}} = - \frac{1}{N} \sum_{j=1}^{N_s} \sum_{k=1}^{N_c} y_{j,k}^i \log(\mathcal{M}^{\text{C}}(x_j^{i}))_k),
\end{equation}
where $\mathcal{L}^{\text{C}}$ denotes the classification loss for the downstream task; $N_c$ denotes the number of classes in the downstream task; $N_s$ denotes the number of available training samples. When classifier training is finished, the performance of the trained classifier model $\mathcal{M}^{\text{C}}$, denoted as $p_n$, will be evaluated and reported to the LWS module for further weight searching, where $n$ denotes the times of training and will be further introduced in the following section.

\section{Low Cost Weight Searching}

The objective of the LWS module is to search for an optimal set of weights (\emph{i.e.}, $\bm{w}^* = \{w^{se}, w^{po}, w^{sp}, w^{pe}\}$), which enables the model to perform optimally across downstream tasks. This is challenging as different downstream tasks require the pre-trained model $\mathcal{M}^0$ to possess an understanding of the semantics of IMU data at various levels. What makes this problem even more difficult is that the relationship between downstream tasks and pre-training tasks is not intuitive. A naive approach is to conduct a grid search over all possible weights $\bm{w}$ to identify the optimal configuration. However, considering that the weights $\bm{w}$ are continuous, and each search requires a complete training cycle that has substantial search costs, the overhead of grid searching becomes impractical in this context. In fact, such a combinatorial optimization problem can be proved to be NP-hard \cite{bartunov2020continuous}. To address this problem, we design a low cost weight searching method based on Bayesian Optimization, which accelerates the search for optimal parameters by learning the relationships between existing weights and their corresponding performances. The core idea of LWS is to utilize a model to learn the relationship of weights and performance and iteratively choose the best weights in the view of the current model. The detail of LWS is shown in Alg. \ref{alg:search} and is introduced below.

\begin{algorithm}[]
%	\SetAlgoNlRelativeSize{0}
	\caption{Low cost Weight Searching based on Bayesian Optimization}
	\label{alg:search}
	\KwIn{initial random weights set $\bm{W}_{\text{ran.}}$, searching budget $N^{\text{bud.}}$, all possible weights $\mathbb{W}$}
	\KwOut{searched optimal weights $\bm{w}_{\text{opt.}}$}
	
	$\bm{W}_{\text{all}} \leftarrow \bm{W}_{\text{ran.}}$ // assign the initial random weights as all searched weights for initialization\;
	$P_{\text{ran.}} \leftarrow$ performance of the model trained with $\bm{W}_{\text{all}}$ on downstream tasks\;
    $P_{\text{all}} \leftarrow P_{\text{ran.}}$ // assign the performance of initial random weights as all performances for initialization\;
	
	\For{$i \leftarrow 1$ \KwTo $N^{\text{bud.}}$}{
		Train a performance model $\mathcal{M}^{\text{P}}$ with $\bm{W}_{\text{all}}$ and $P_{\text{all}}$ // train a new $\mathcal{M}^{\text{P}}$ each time before searching\;
        $E \leftarrow \{\}$ // initialize the set of expected improvement (EI) as an empty set\;
        \For{$\bm{w}_j \in \mathbb{W}$}{
            $\epsilon_j \leftarrow$ calculated expected improvements of weights $\bm{w}_j$ using Equation \ref{eq:ei}\;
            $E \leftarrow E \cup \{\epsilon_j\}$ // add the EI of weight $\bm{w}_j$ to the EI set\;
        }
		$\bm{w}_{\text{new}} \leftarrow \arg_{\bm{w}_j} \max_{\epsilon_j} E$ // assign the weights with best-predicted performance as the new trial weights\;
		$p_{\text{new}} \leftarrow$ performance of the model after pre-training with $\bm{w}_{\text{new}}$ and fine-tuning on downstream tasks\;
		$\bm{W}_{\text{all}} \leftarrow \bm{W}_{\text{all}} \cup \bm{w}_{\text{new}}$ // add the weights in this loop to weights set for the training of next loop\;
		$P_{\text{all}} \leftarrow P_{\text{all}} \cup p_{\text{new}}$ add the performance in this loop to the performance set for the training of the next loop\;
	}
	$\bm{x}_{\text{opt.}} \leftarrow \arg_{\bm{w}_j} \min_{p_j} P_{\text{all}}$ // return the weights with the best evaluated performance\;
\end{algorithm}

\subsection{Weight-Performance Learning}

Given all possible weights, denoted as $\mathbb{W}$, we first randomly sample an initial set of several weights in $\mathbb{W}$, denoted as $\bm{W}_{\text{ran.}}$ and their corresponding evaluated model performances denoted as $P_{\text{ran.}}$. A performance model is employed to model the unknown relationship between weights, denoted as $\bm{w}_k$, and their corresponding performances, denoted as $p_k$. In this paper, we choose the Gaussian Process model for its high efficiency and superior performance, which can be defined as follows \cite{schulz2018tutorial}: $\mathcal{M}^{\text{P}}(\bm{w}) \sim \mathcal{N}(\mu(\bm{w}), k(\bm{w}, \bm{w}'))$,
where $\mu(\cdot)$ denotes the mean function and $k(\bm{w}, \bm{w}'))$ denotes the covariance function of all pairs of weights $ (\bm{w} \in \mathbb{W}, \bm{w}' \in \mathbb{W})$. For each weight $\bm{w}_i$, a prediction of performance, denoted as $p_j^{\text{inf.}}$, and the corresponding uncertainty, denoted as $c_j^{\text{inf.}}$, can be predicted by the Gaussian Process model $\mathcal{M}^{\text{P}}$, \emph{i.e.}, $p_j^{\text{i.}}, c_j^{\text{i.}} = \mathcal{M}^{\text{P}}(\bm{w}_i)$.

\subsection{Best Weights Searching}

For the acquisition of best weights, we choose the Expected Improvement (EI) algorithm measuring the potential of weights $\bm{w}_i$ to improve upon the current best performance, denoted as $p^{\text{best}}$. Specifically, the Expected Improvement of all possible weights, denoted as $\epsilon_i$ for weights $\bm{w}_i$, is calculated as follows \cite{qin2017improving}:
\begin{equation}
    \label{eq:ei}
    \begin{aligned}
    \epsilon_i &= \mathbb{E}[\max (0, \mathcal{M}^{\text{P}}(\bm{w}_i) - p^{\text{best}})] \\
    &= (p_j^{\text{i.}} - p^{\text{best}}) \Phi(\frac{p_j^{\text{i.}} - p^{\text{best}}}{c_j^{\text{i.}}}) + c_j^{\text{i.}} \phi(\frac{p_j^{\text{i.}} - p^{\text{best}}}{c_j^{\text{i.}}}),
    \end{aligned}
\end{equation}
where $\mathbb{E}[\cdot]$ denotes the mathematical expectation function; $\Phi(\cdot)$ and $\phi(\cdot)$ denotes the cumulative distribution function (CDF) and probability density function (PDF), respectively. Note that EI takes into account both the predicted value and uncertainty of model $\mathcal{M}^{\text{P}}$ to estimate the potential improvement in the objective function that might result from sampling at weights $\bm{w}_i$, where the first term (\emph{i.e.}, $(p_j^{\text{i.}} - p^{\text{best}}) \Phi(\frac{p_j^{\text{i.}} - p^{\text{best}}}{c_j^{\text{i.}}})$) represents the expected improvement when the predicted value is actually better than the current optimal value $p^{\text{best}}$ and the second term (\emph{i.e.}, $c_j^{\text{i.}} \phi(\frac{p_j^{\text{i.}} - p^{\text{best}}}{c_j^{\text{i.}}})$) considers the uncertainty of the prediction, and calculates the expected improvement under this uncertainty.
The set of expected improvement of all possible weights $\mathbb{W}$ is denoted as $E$. The best weights with highest expected improvement in $E$ among all possible weights $\mathbb{W}$, denoted as $\bm{w}_{\text{new}}$, will be selected for the training and evaluation in the current loop, \emph{i.e.}, $\bm{w}_{\text{new}} = \arg_{\bm{w}_j} \max_{\epsilon_j} E$. In this loop, the model is first pre-trained with weights $\bm{w}_{\text{new}}$ and then fine-tuned on downstream tasks to obtain the corresponding performance $p_{\text{new}}$. $\bm{w}_{\text{new}}$ and $p_{\text{new}}$ will then be added to the historical weights and performances, denoted as $\bm{W}_{\text{all}}$ and $P_{\text{all}}$, respectively, \emph{i.e.}, $\bm{W}_{\text{all}} = \bm{W}_{\text{all}} \cup \bm{w}_{\text{new}}, P_{\text{all}} = P_{\text{all}} \cup p_{\text{new}}$. The updated $\bm{W}_{\text{all}}$ and $P_{\text{all}}$ will be utilized for the training of $\mathcal{M}^{\text{P}}$ in the next loop. The optimization loop iterates until the searching budget is exhausted or the validation outcomes converge.

\begin{table}[]
	\caption{Saga is implemented on 5 different types of mobile phones with distinct hardware configurations.}
	\label{tab:hardware}
	\centering
	\scalebox{1}{
		\begin{tabular}{c c c c}
			\toprule
			Phone              & SoC            & Memory & Disk  \\ \hline
			Mi 6               & Snapdragon 835 & 6GB    & 64GB  \\
			Pixel 3 XL         & Snapdragon 845 & 4GB    & 128GB \\
			Honor v9           & Kirin 960      & 6GB    & 64GB \\
			Mi 10              & Snapdragon 870 & 6GB    & 128GB \\
			Mi 11              & Snapdragon 888 & 8GB    & 256GB \\
			\bottomrule
		\end{tabular}
	}
   % \vspace{-0.3cm}
\end{table}

\section{Evaluation}

\subsection{Methodology}

\subsubsection{Implementation}

We implement Saga on an Ubuntu server equipped with 256 GB of memory and 4 Nvidia 3090 GPUs. Our implementation is based on LIMU \cite{xu2021limu} and incorporates multi-level masking techniques and weight searching. We utilize Pytorch for neural network training due to its widespread adoption and flexibility in deep learning applications. For IMU signal processing, we rely on Scipy, a powerful scientific computing library. Additionally, the GP model employed for weight searching is realized using Scikit-learn, a machine-learning library providing various tools for data mining and data analysis.

The pre-training model we adopt is BERT \cite{joshi2020spanbert,xu2021limu}, which comprises 4 lightweight transformer blocks, each featuring a hidden dimension of 72. This model has proven effective in various NLP tasks. For the downstream task, we opt for a GRU classifier, as it has demonstrated superior performance in classification tasks according to \cite{xu2021limu}. We train our system using the Adam optimizer with a learning rate set to 1e-3. The training process consists of two phases: an initial pre-training for 50 epochs on all unlabelled data, followed by fine-tuning for another 50 epochs on a small amount of labelled data. All parameters are kept trainable during fine-tuning for better performance \cite{xu2021limu,yosinski2014transferable}.

\begin{table}[]
	\caption{Datasets summary (A=accelerometer, G=gyroscope, M=magnetometer).}
	\label{tab:dataset}
        \centering
	\scalebox{0.95}{
	\begin{tabular}{ccccccc}
		\toprule
		Dataset & Sensor  & Activity & User & Placement & Window & Sample \\ \hline
		HHAR    & A, G    & 6        & 9    & -    & 120  & 9166   \\ 
		Motion  & A, G    & 6        & 24   & -   & 120        & 4534   \\ 
		Shoaib  & A, G, M   & 7        & 10   & 5   & 120        & 10500   \\
		% SHL     & A, G, M & 24       & 3    & 4   & 500       & 12100 \\ 
  \bottomrule
	\end{tabular}
	}
% \vspace{-0.3cm}
\end{table}

\begin{table}[]
	\caption{Summary of tasks considered for evaluation.}
	\label{tab:task}
	\centering
	\scalebox{0.95}{
		\begin{tabular}{cccc}
			\toprule
			Task & Description          & Labels                        & Datasets                       \\ \hline
			AR   & activity recognition & walk, run, etc.      & HHAR, Motion \\
			UA   & user authentication  & user 1, user 2, etc.        & HHAR, Shoaib \\
			DP   & device positioning     & hand, torso, etc.           &  Shoaib              \\
			% DT   & device type recognition     & Samsung, LG, etc.           & HHAR              \\  
   \bottomrule
		\end{tabular}
	}
% \vspace{-0.3cm}
\end{table}

To evaluate the performance of Saga in real-world scenarios, we have implemented it on 5 representative mobile phones: Mi 6, Pixel 3 XL, Honor v9, Mi 10, and Mi 11. The specific hardware configurations of these devices are detailed in Table \ref{tab:hardware}. Finally, the trained models are deployed on these mobile phones using ONNX Runtime \cite{onnxruntime}, which provides an efficient runtime for deep learning models.

\subsubsection{Datasets}

\begin{figure*}
	\centering
	\subfigure[labelling rate = 5\%]{
		\label{fig:label_0.05}
		\centering
		\begin{minipage}{0.23\textwidth}
			\includegraphics[width=\linewidth]{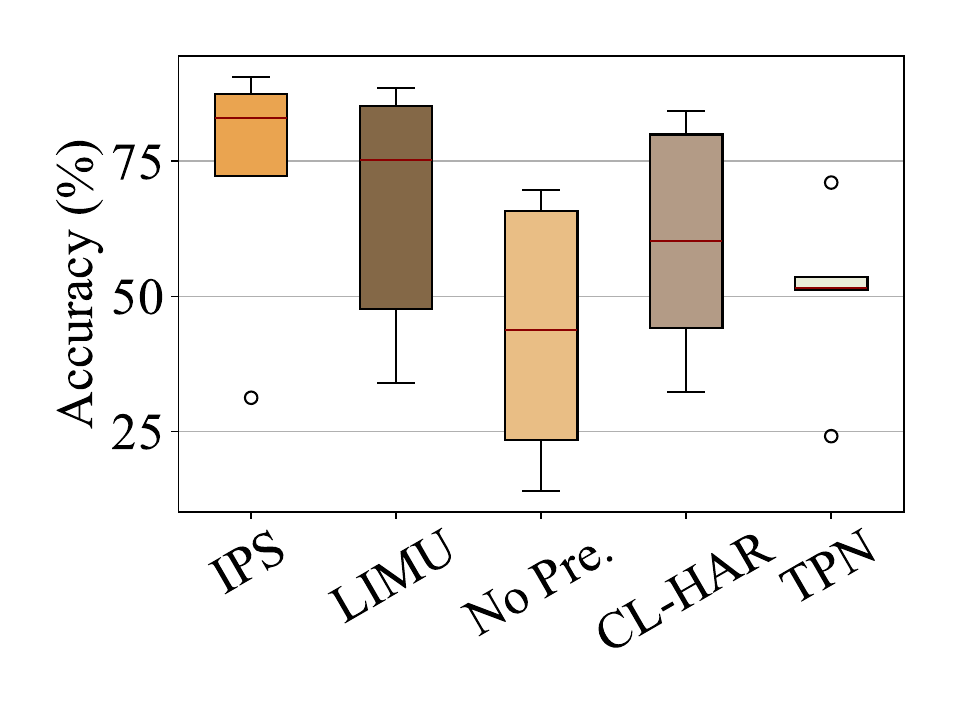}
			\includegraphics[width=\linewidth]{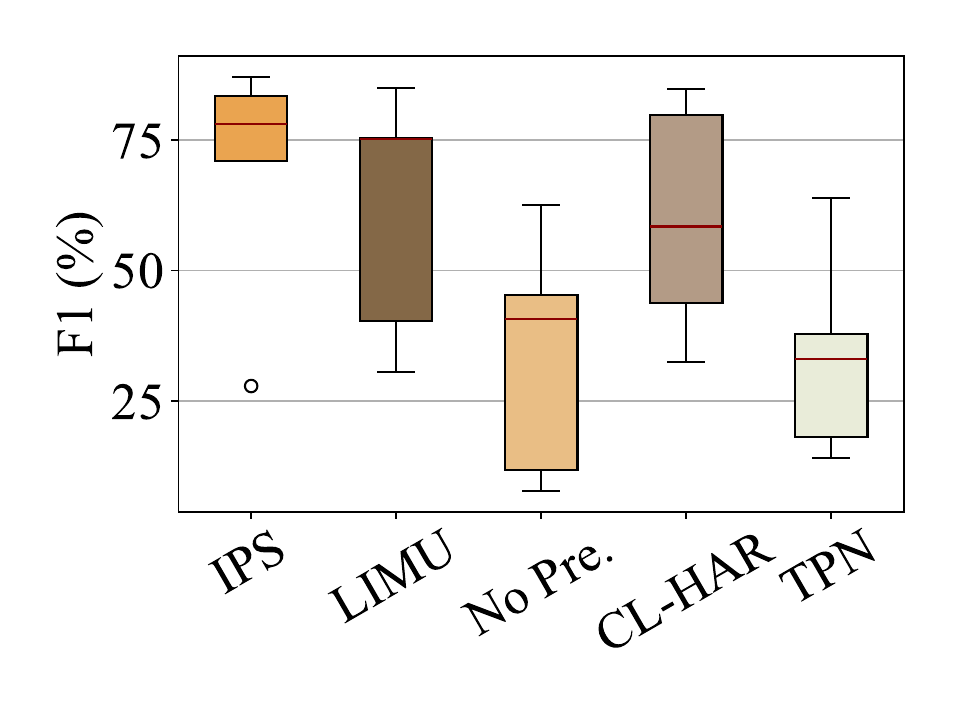}
		\end{minipage}
	}
	\subfigure[labelling rate = 10\%]{
		\label{fig:label_0.1}
		\centering
		\begin{minipage}{0.23\textwidth}
			\includegraphics[width=\linewidth]{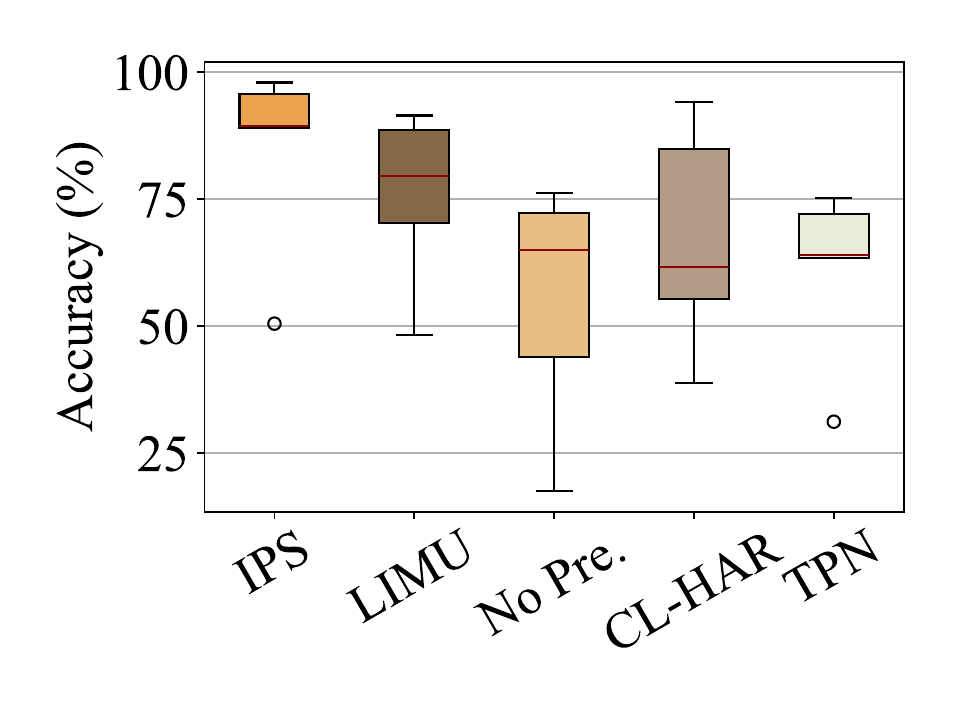}
			\includegraphics[width=\linewidth]{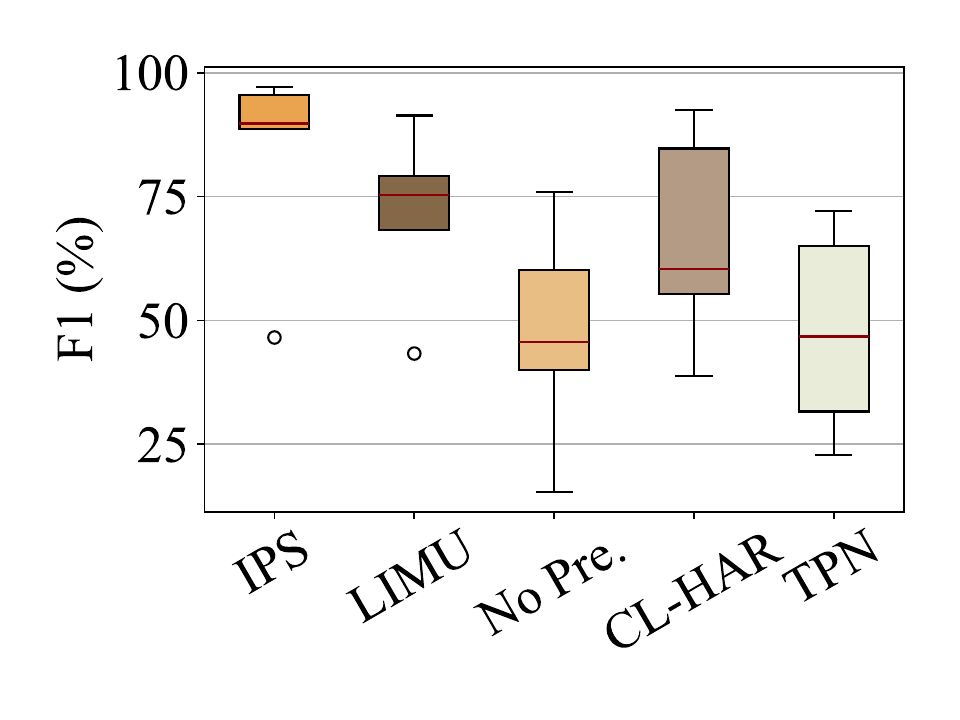}
		\end{minipage}
	}
	\subfigure[labelling rate = 15\%]{
		\label{fig:label_0.15}
		\centering
		\begin{minipage}{0.23\textwidth}
			\includegraphics[width=\linewidth]{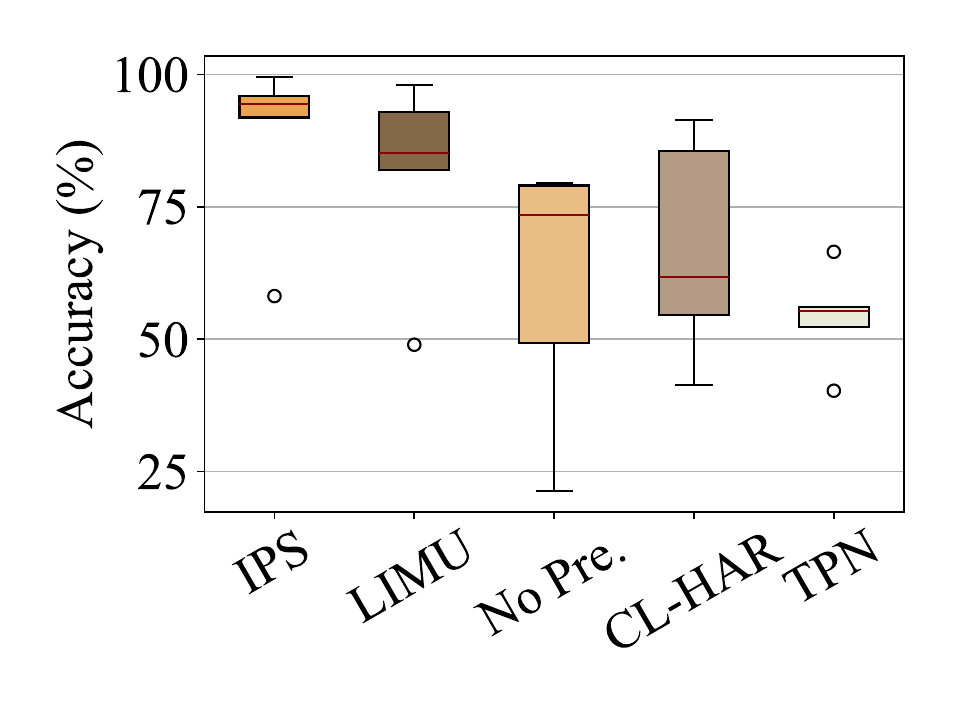}
			\includegraphics[width=\linewidth]{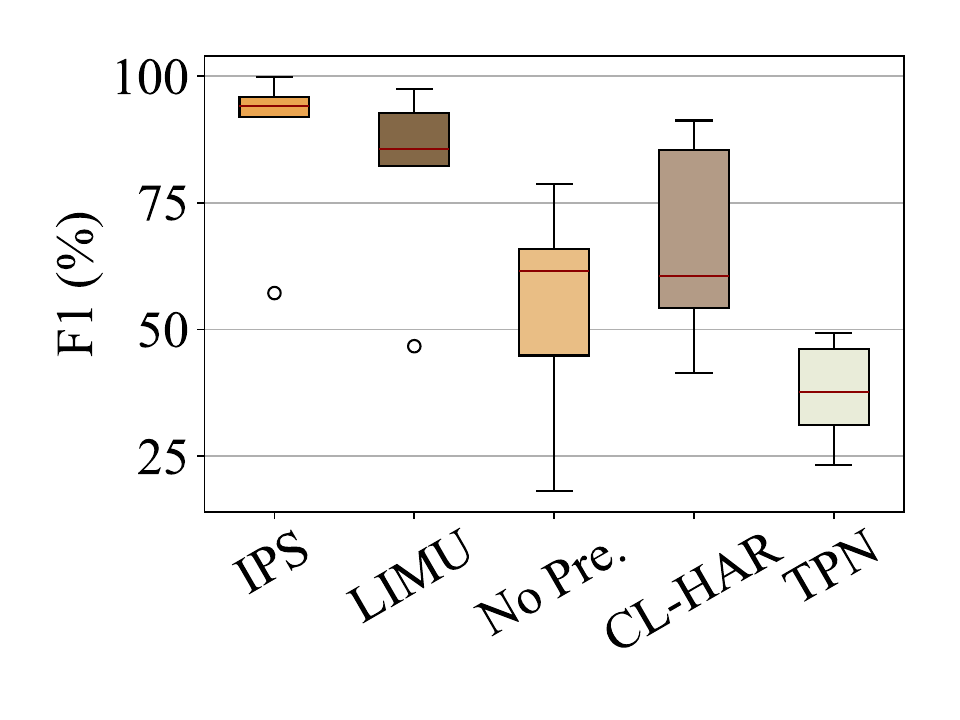}
		\end{minipage}
	}
	\subfigure[labelling rate = 20\%]{
		\label{fig:label_0.2}
		\centering
		\begin{minipage}{0.23\textwidth}
			\includegraphics[width=\linewidth]{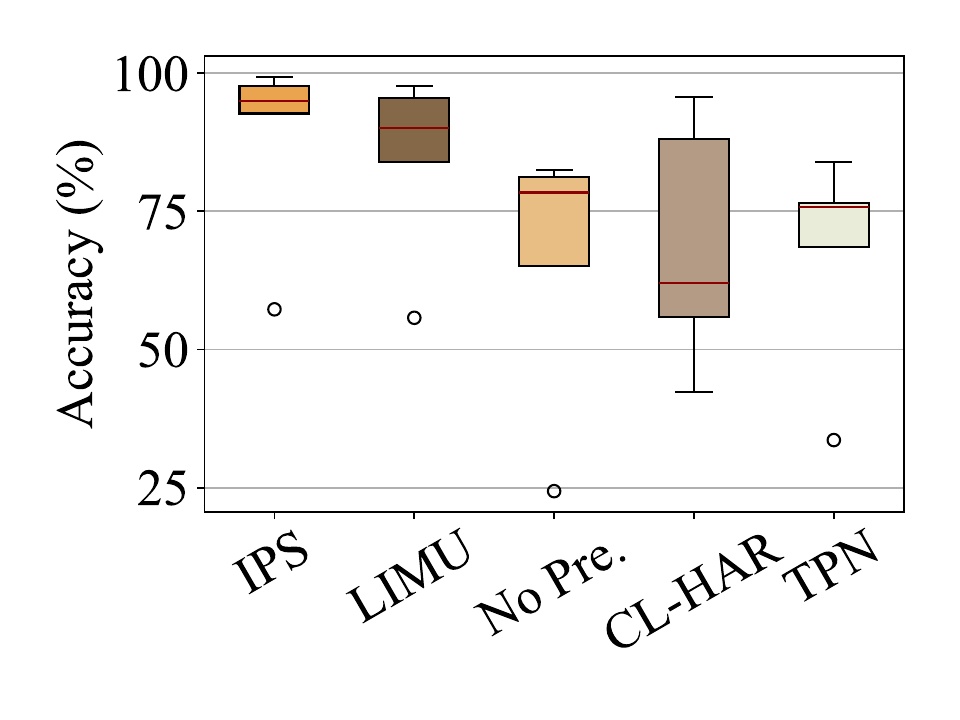}
			\includegraphics[width=\linewidth]{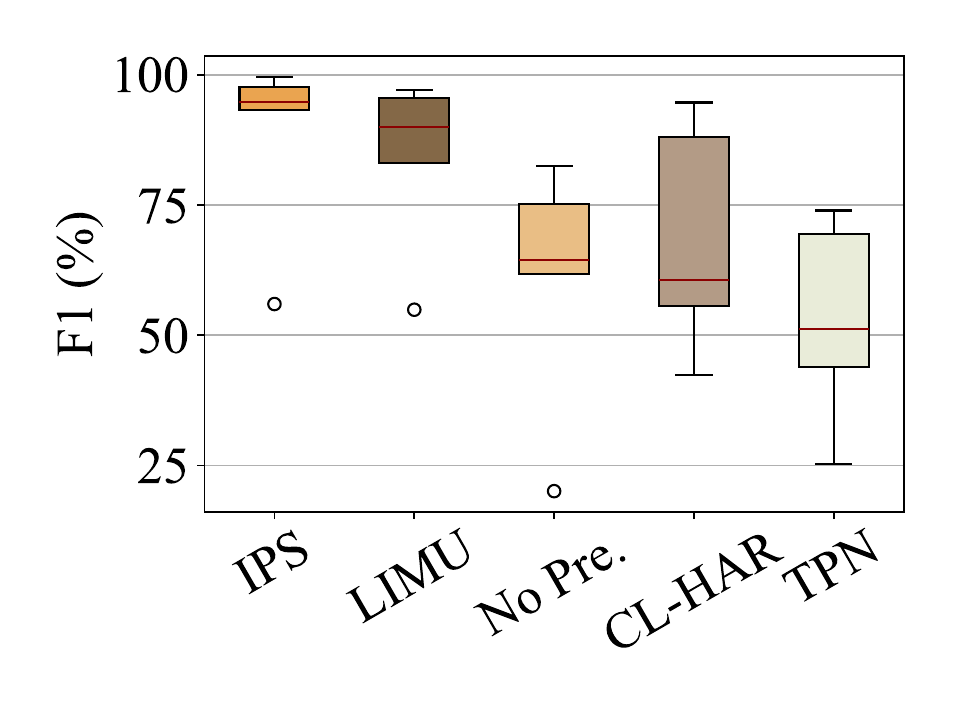}
		\end{minipage}
	}
   % \vspace{-0.2cm}
	\caption{Boxplot of the perception of Saga and other candidate methods on all three tasks on all three datasets with various labelling rates, \emph{i.e.}, 5\%, 10\%, 15\%, and 20\%, where Saga can outperform all other methods.}
   % \vspace{-0.5cm}
	\label{fig:boxplot}
\end{figure*}

We consider the following four user perception datasets, which have been commonly used in previous works \cite{xu2021limu, qian2022makes}, as summarized in Table \ref{tab:dataset}.

\begin{itemize}
	\item \textbf{HHAR} \cite{stisen2015smart}: The HHAR dataset is a publicly available dataset consisting of accelerometer and gyroscope readings collected from 6 types of mobile phones (3 models of Samsung Galaxy and 1 model of LG). The smartphones are worn around the waist by 9 users performing 6 different activities, with sampling rates in 100 - 200 Hz.
	% \item \textbf{UCI} \cite{reyes2016transition}: The UCI dataset contains accelerometer and gyroscope readings from a Samsung Galaxy S II smartphone carried by 30 subjects when performing six activities, \emph{i.e.}, standing, sitting, lying, walking, going downstairs, and going upstairs. The data sampling rate is 50 Hz.
	\item \textbf{Motion} \cite{malekzadeh2019mobile}: The Motion dataset is a publicly available dataset of accelerometer and gyroscope readings collected from a smartphone (iPhone 6s) worn by 24 subjects during various daily activities. The data is collected with the smartphone in the front pockets of the subjects. Motion covers 6 different activities. Motion includes accelerator and gyroscope data sampled at 50 Hz.
	\item \textbf{Shoaib} \cite{shoaib2014fusion}: The Shoaib dataset gathered data pertaining to seven distinct physical activities, namely walking, sitting, standing, jogging, biking, walking upstairs, and walking downstairs. In the course of this data acquisition, ten male volunteers participated, each equipped with five Samsung Galaxy SII (i9100) smartphones strategically positioned at five different body locations: right pocket, left pocket, belt, upper arm, and wrist. These smartphones recorded accelerometer, gyroscope, and magnetometer readings at a frequency of 50 samples per second.
	% \item \textbf{SHL} \cite{wang2021three}: The data collection campaign of the SHL dataset employed three full-time participants. Four HUAWEI Mate 9 smartphones were respectively placed on four different body locations of a participant, including hand, torso, backpack, and trousers' front pocket. A data logging application \cite{ciliberto2017high} was used to automatically log 16 sensor modalities including IMU sensors at a sampling rate of 100 Hz.
\end{itemize}

For HHAR, Shoaib, and Motion datasets, we first down-sample the IMU samples to 20 Hz and slice the IMU samples with a window of 6s (120 data points). 
All data samples are normalized as follows:
$a^*_k = \frac{a_k}{\mathit{g}}, m^*_k = {m_k} / {\sqrt{\sum{m_k^2}}}, \text{for } k \in \{x, y, z\}$,
where $a_k$ and $m_k$ denote the values of accelerometers and magnetometers on the coordinate axis $k$, respectively; $\mathit{g}$ denotes the universal gravitational constant. After pre-processing, the HHAR, Motion, and Shoaib datasets consist of 9,166, 4,534, and 10,500 samples, respectively.
All datasets are divided into training sets, validation sets, and testing sets with a ratio of 6:2:2.

%Figure \ref{fig:semantics}(b) illustrates the paradigm for deep learning on IMU data. IMU data is continuous over time. However, neural networks can only accept fixed-length inputs. As a result, to be processed by neural network, we have to preprocess IMU data into a fixed length (in another word, a fixed recording time for a fixed IMU sampling rate). 
%A commom practice for IMU partition is to partition IMU data with a fixed window. The length of the window is pre-set based on prior knowledge. For example, LIMU \cite{xu2021limu} chooses to partition the IMU data by a fixed window of 6 seconds, considering that human actions do not last for more than 6 seconds under normal circumstances \cite{reyes2016transition}. 

\subsubsection{Candidate Methods}

We consider the following three candidate methods:
\begin{itemize}
	\item \textbf{LIMU} \cite{xu2021limu}: LIMU pre-trains on IMU data by masking the IMU data at a level of data point. The model is pre-trained by reconstructing the masked IMU sample.
	\item \textbf{CL-HAR} \cite{qian2022makes}: CL-HAR pre-trains on IMU data by contrastive learning. A single IMU sample is first transformed into a group of views and the model is pre-trained by distinguishing the different transforms of a single sample.
	\item \textbf{TPN} \cite{saeed2019multi}: TPN pre-trains on IMU data by classifying between different transforms. The model is pre-trained to distinguish different transforms on IMU data.
\end{itemize}

For the transforms used in CL-HAR and TPN, we choose to use complete data augmentation \cite{xu2023practically} (\emph{i.e.}, the augmentation function which can be fully formulated with original observations and known physical states) on IMU data as transforms for its better performance on IMU data.

\subsubsection{Tasks and Metrics}

We consider three typical user perception tasks as shown in Table \ref{tab:task}. All these tasks belong to the classification task. Therefore, we adopt accuracy (Acc) and F1 score (F1) for performance comparison. Acc is defined as the proportion of correctly predicted samples to the total number of test samples and F1 is defined as $\text{F1}=\frac{1}{N_c}\sum_{i=1}^{N_c}\frac{2 p_i r_i}{p_i + r_i}$, where $p_i$ and $r_i$ denote the precision and recall of the $i$-th class, respceptively, and $N_c$ denotes the number of all classes.

\subsection{Overall IUP Accuracy}
\label{sec:perf}

We now illustrate the efficacy of Saga across various tasks characterized by low labelling rates. To assess its performance, we compare the perception accuracy of Saga against other leading methods, namely LIMU, CL-HAR, and TPN, at labelling rates of 5\%, 10\%, 15\%, and 20\%. Furthermore, to underscore the value of pre-training, we contrast Saga with an approach that forgoes pre-training and relies solely on labelled data for training, neglecting unlabelled data. Figure \ref{fig:boxplot} presents the average relative accuracy and F1 score (relative to the SOTA method, \emph{i.e.}, LIMU, when trained with all labelled data samples) on all tasks and datasets of the different methods across labelling rates of 5\%, 10\%, 15\% and 20\%.

\textbf{Pre-training can improve user perception accuracy when only a few labelled samples are provided.}
It can be observed that, apart from TPN (which often fails to converge in many cases), the methods employing pre-training demonstrate superior performance compared to those without pre-training. For example, Saga and LIMU outperform the naive method without pre-training on average by over 30\% in terms of prediction accuracy.
This underscores the effectiveness of the pre-training approach. 

\textbf{Generative pre-training performs better than contrastive-learning-based pre-training on IMU data.} 
We can see that masking-based pre-training methods (\emph{i.e.}, LIMU and Saga) always outperform contrastive-learning-based pre-training methods (\emph{i.e.}, CL-HAR and TPN). For example, Saga and LIMU outperform CL-HAR and TPN on average by over 15\% and 35\% in terms of prediction accuracy, respectively. This is because of the difficulty of generating effective views for IMU pre-training.

% \begin{figure*}
% 	\centering
% 	\includegraphics[width=0.94\linewidth]{figures/bar_hhar_013.pdf}
% 	\includegraphics[width=0.94\linewidth]{figures/bar_hhar2_motion3.pdf}
% 	\includegraphics[width=0.94\linewidth]{figures/bar_shoaib3.pdf}
% 	\includegraphics[width=0.94\linewidth]{figures/bar_shl3.pdf}
% 	\caption{Detailed results of top-3 candidate methods on all 8 tasks with labelling rates of 5\%, 10\%, 15\% and 20\%, where Saga significantly outperforms other candidate methods.}
% 	\label{fig:results}
% \end{figure*}

\begin{figure}
    \centering
    \includegraphics[width=\linewidth]{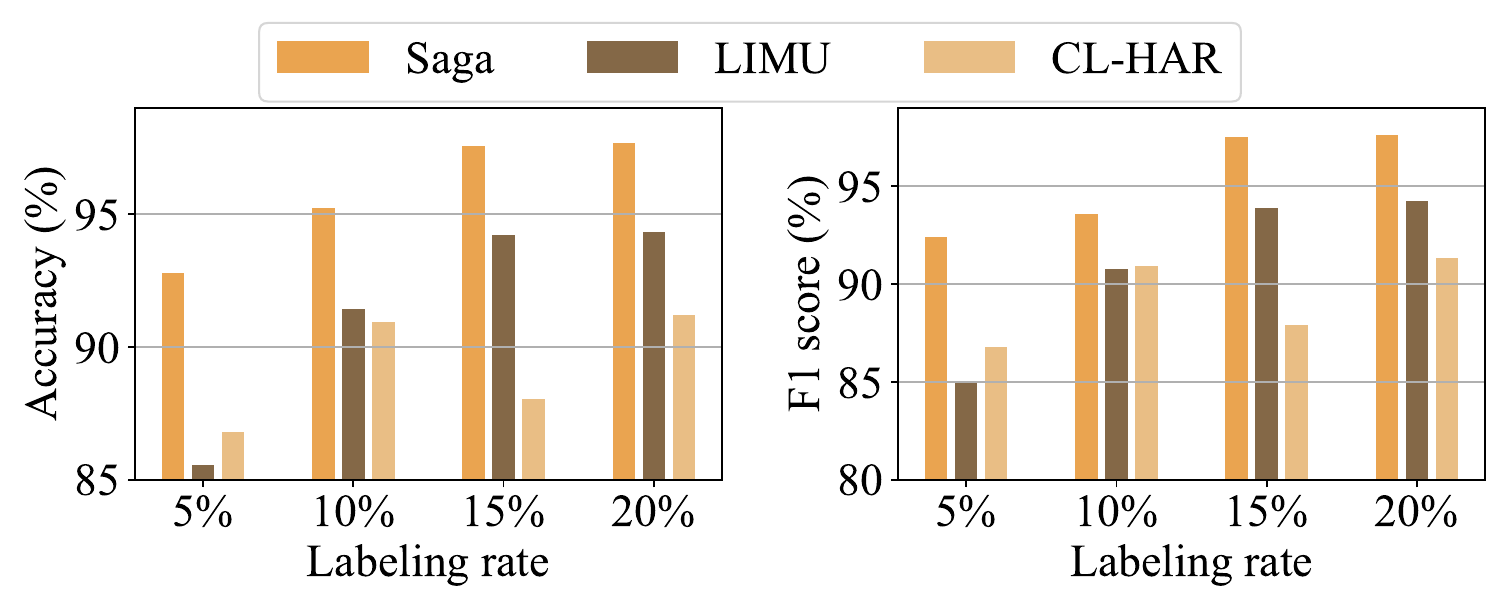}
    \caption{Detailed results of top-3 candidate methods on AR task on HHAR dataset with labelling rates of 5\%, 10\%, 15\% and 20\%, where Saga significantly outperforms other candidate methods.}
    \label{fig:ar-hhar}
\end{figure}

\textbf{Saga pre-training based on IMU semantics outperforms SOTA methods.}
Although the performance of masking-based methods is better than other methods, Saga can always achieve the best performance within the masking-based category. For example, Saga can outperform LIMU on average by around 10\% in terms of prediction accuracy when labelling rate is larger than 10\% (around 100 samples per class), with a relative accuracy of over 90\% on average. This is because Saga can extract semantics specific to downstream tasks, making feature extraction more accurate and effective. 

\begin{figure}
	\centering
	\includegraphics[width=\linewidth]{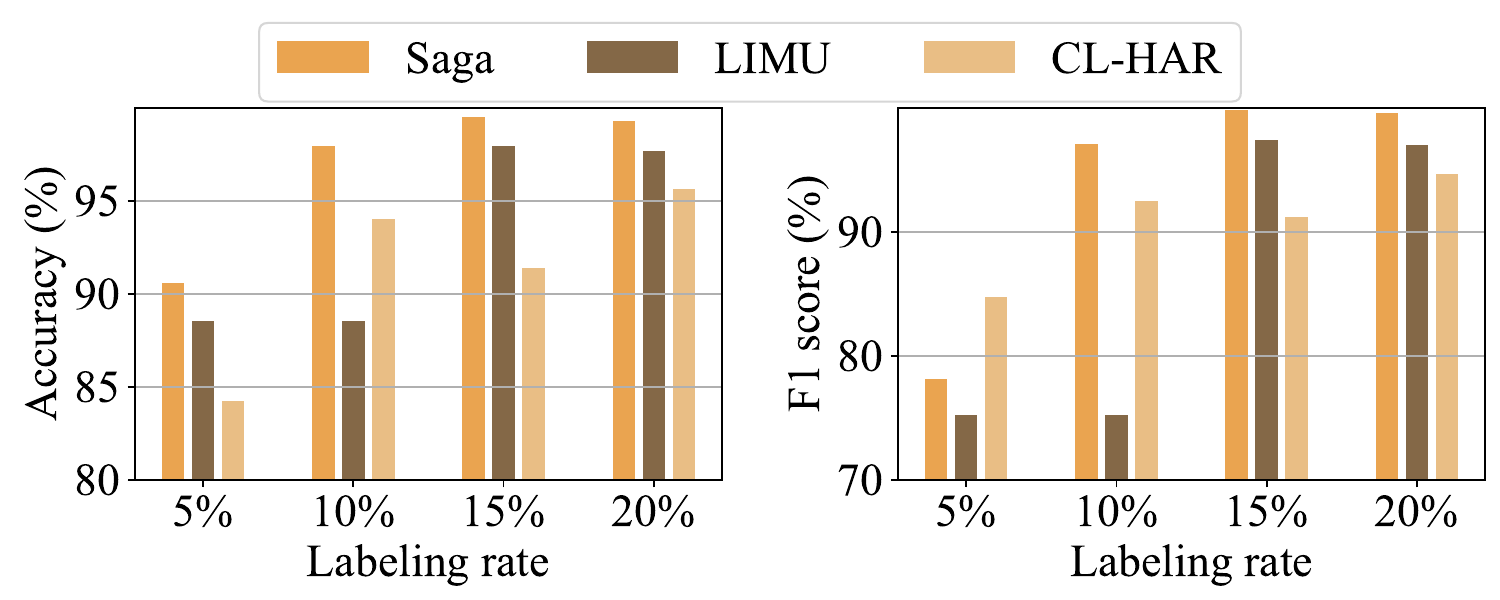}
	\caption{Detailed results of top-3 candidate methods on AR task on Motion dataset with labelling rates of 5\%, 10\%, 15\% and 20\%, where Saga significantly outperforms other candidate methods.}
	\label{fig:ar-motion}
\end{figure}

\begin{figure}
    \centering
    \includegraphics[width=\linewidth]{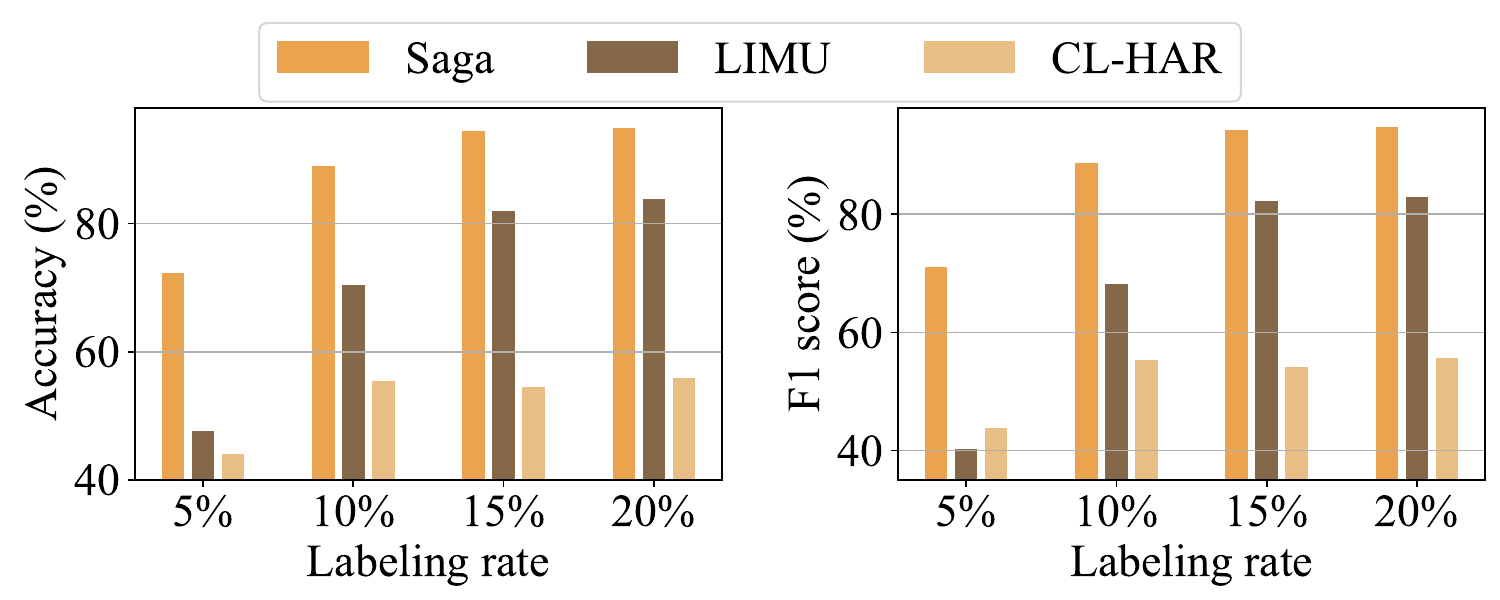}
    \caption{Detailed results of top-3 candidate methods on UA task on HHAR dataset with labelling rates of 5\%, 10\%, 15\% and 20\%, where Saga significantly outperforms other candidate methods.}
    \label{fig:ua-hhar}
\end{figure}

\begin{figure}
    \centering
    \includegraphics[width=\linewidth]{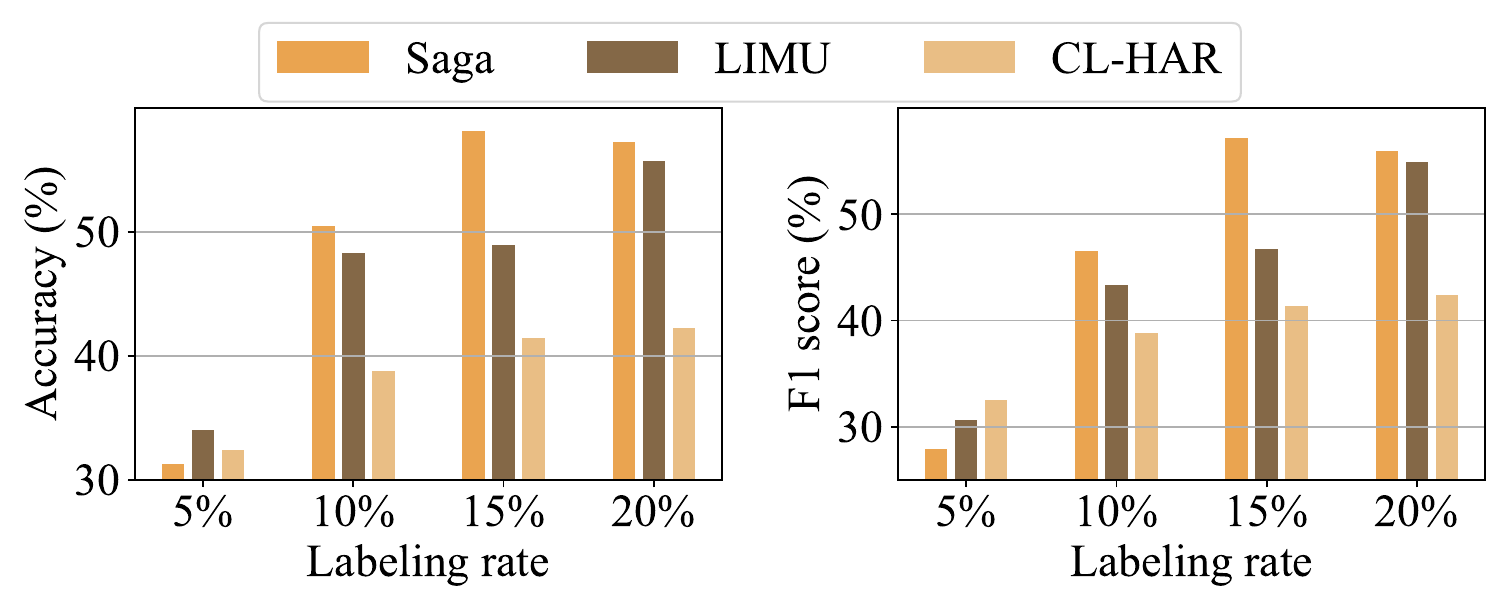}
    \caption{Detailed results of top-3 candidate methods on UA task on Shoaib dataset with labelling rates of 5\%, 10\%, 15\% and 20\%, where Saga significantly outperforms other candidate methods.}
    \label{fig:ua-shoaib}
\end{figure}

\begin{figure}
	\centering
	\includegraphics[width=\linewidth]{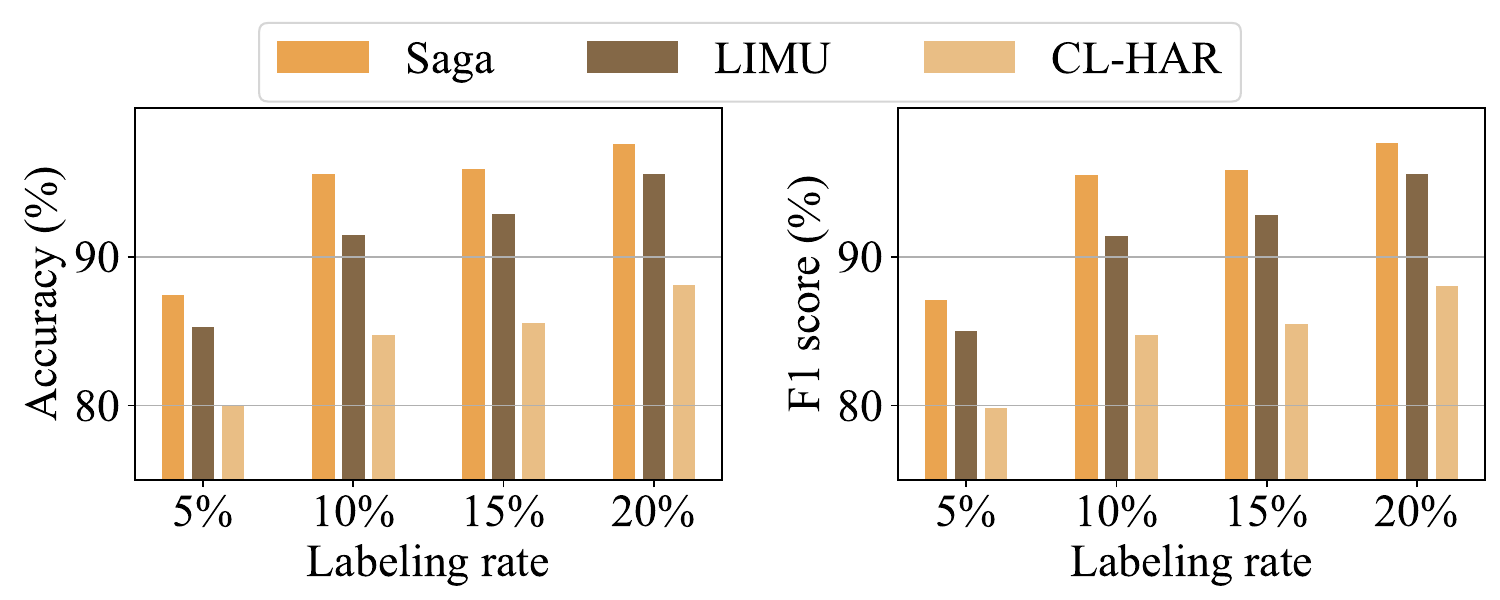}
	\caption{Detailed results of top-3 candidate methods on DP task on Shoaib dataset with labelling rates of 5\%, 10\%, 15\% and 20\%, where Saga significantly outperforms other candidate methods.}
	\label{fig:dp-shoaib}
\end{figure}

For a better understanding of the performance of different candidate methods, we present the performance of top-3 candidate methods (\emph{i.e.}, Saga, LIMU, and CL-HAR) on all three tasks on all three datasets with four labelling rates in Fig. \ref{fig:ua-hhar}, Fig. \ref{fig:ar-hhar}, Fig. \ref{fig:ar-motion}, Fig. \ref{fig:dp-shoaib} and Fig. \ref{fig:ua-shoaib}, respectively. 

\textbf{Saga performs extremely well even under extremely low labelling rates.} We can see that Saga often achieves significant performance improvement when the labelling rate is low. For example, for the UA task on the HHAR dataset, when the labelling rate is 5\%, Saga achieves over 20\% (up to more than 50\% in some tasks, \emph{e.g.}, 51.6\% in terms of perception accuracy on UA task from HHAR dataset) improvement compared to LIMU and CL-HAR in terms of both accuracy and F1 score. When only 80 labelled samples are used, Saga can outperform LIMU by 11.8\%. This is because Saga better extracts the unsupervised semantics of IMU data, enabling it to learn more generalizable features from less data. We notice that as the labelling rate increases, the accuracy difference between different methods gradually narrows. This is because sufficient labelled data reduces the model's reliance on pre-training.

\subsection{Ablation Experiments}

\begin{figure}[]
	\centering
	\subfigure[Accuracy]{
		\label{fig:masking_acc}
		\centering
		\begin{minipage}[]{0.22\textwidth}
			\includegraphics[width=1.1\linewidth]{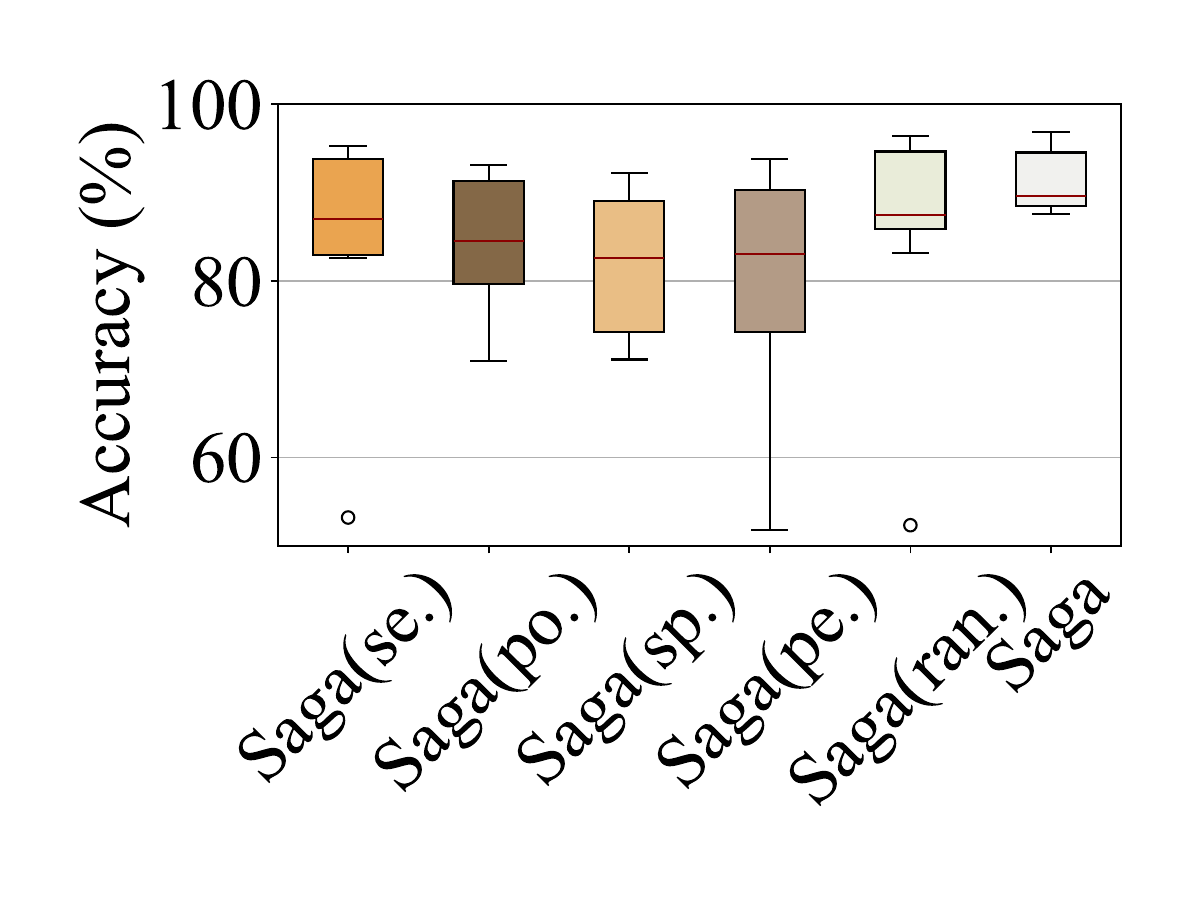}
		\end{minipage}
	}
	\subfigure[F1 score]{
		\label{fig:masking_f1}
		\centering
		\begin{minipage}[]{0.22\textwidth}
			\includegraphics[width=1.1\linewidth]{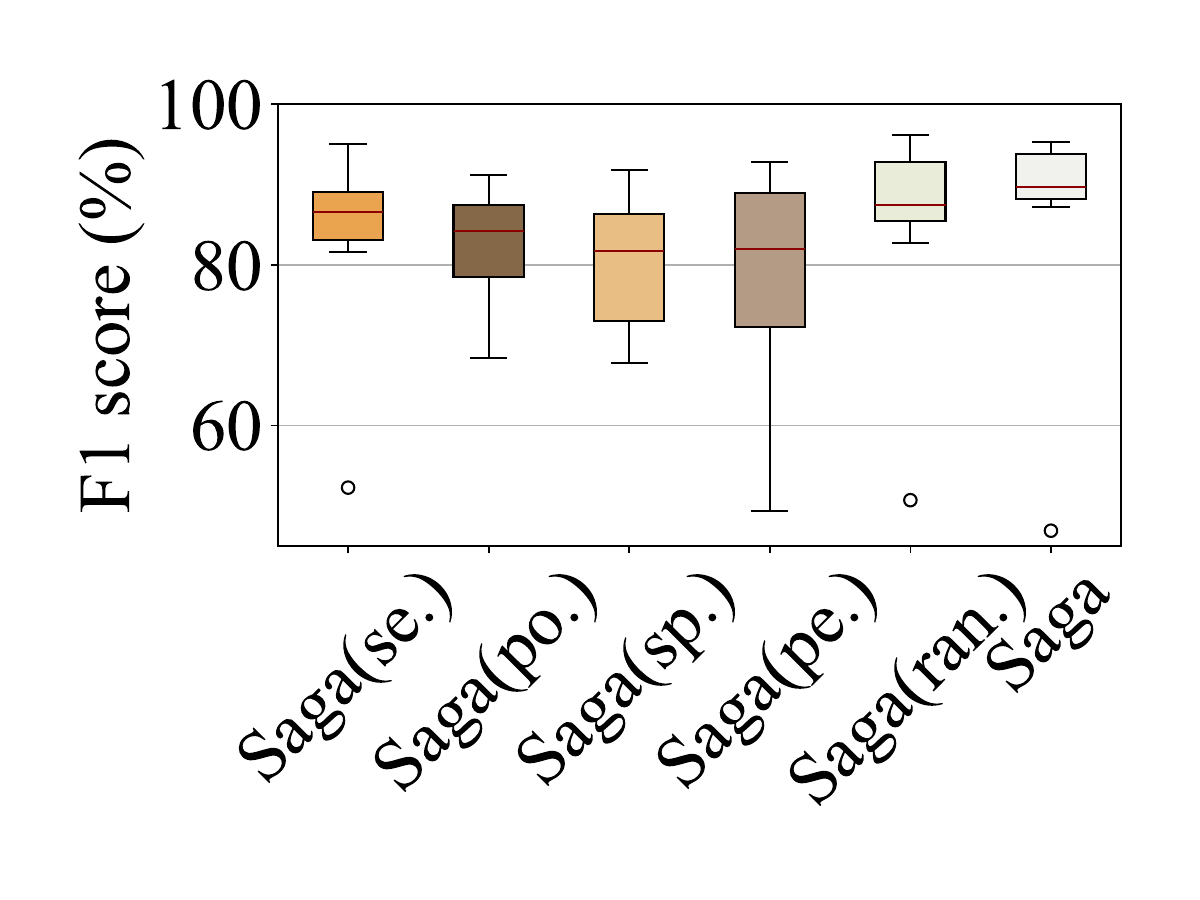}
		\end{minipage}
	}
	% \vspace{-0.2cm}
	\caption{Boxplot of average relative accuracy and F1 score of all masking tasks for pre-training with labelling rates of 5\%, 10\%, 15\% and 20\%.}
	% \vspace{-0.6cm}
	\label{fig:boxplot_masking}
\end{figure}

To show the effectiveness of each individual semantic-based masking method, we pre-train the model only by sensor-level masking, sub-period-level masking, and period-level masking individually, and then test them when fine-tuned on different tasks, denoted as \emph{Saga (se.)}, \emph{Saga (po.)}, \emph{Saga (sp.)} and \emph{Saga (pe.)}, respectively. Moreover, to show the effectiveness of our weight searching method, we compare it with the results of a set of random weights, and denote the results as \emph{Saga (ran.)}. Figure \ref{fig:boxplot_masking} presents the average relative performance of all masking tasks with varying labelling rates of 5\%, 10\%, 15\%, and 20\%.

\textbf{Our proposed masks are as effective as point-level masking used in LIMU.} LIMU works well on AR task, which only utilizes point-level masking, \emph{i.e.}, Saga(po.). We can see from Figure \ref{fig:boxplot_masking} that the masks of the other three levels work as well as Saga(po.). On average, Saga(se.) can outperform Saga(po.) by over 3\% in terms of perception accuracy; The perception accuracy of Saga(sp.), and Saga(pe.) is also very close to Saga(po.). This shows the effectiveness of our proposed masks in Saga.

\textbf{Combination of multiple masks works better than only using one mask.} We can see that Saga(ran.), which utilizes multiple pre-training tasks but with random weights, can outperform all Saga(se.), Saga(po.), Saga(sp.), and Saga(pe.). This is because of the intuitive relationships between downstream tasks and pre-training tasks. One downstream task may benefit from multiple pre-training tasks from different degrees. As a result, a combination of multiple tasks can also outperform the methods just using one pre-training task.

\textbf{Our proposed low cost weight searching (LWS) module can further enhance the user perception accuracy.} Allowing the combination of multiple pre-training tasks as well, Saga always performs better than Saga(ran.), where the weights are randomly selected. On average, Saga can outperform Saga(ran.) by over 5\% in terms of both worst perception accuracy and F1 score, respectively.
This is because of Bayesian Optimization utilized by Saga for weights searching, which models a probability relationship between weights and corresponding performances. As a result, LWS is more directional and therefore more efficient.

\subsection{System Costs}

% Finally, we evaluate the system costs of Saga and all other methods. 

\subsubsection{Training Costs}

We first investigate the training costs, in terms of the train time of one batch containing 32 samples with a window length of 120, parameters in the model, the disk space consumption of the model, and the GPU memory consumption when training with a batch size of 2048. Table \ref{tab:costs} shows the training costs of all candidate methods.

\begin{table}[]
	\caption{Training costs of all candidate methods.}
   % \vspace{-0.2cm}
	\centering
	\label{tab:costs}
	\scalebox{1}{
		\begin{tabular}{ c c c c c }
			\toprule
			Methods        & LIMU & CL-HAR & TPN & Saga \\ \hline
			Train time (ms)    & 31 & 35 & 7 & 56  \\
			Parameters (KB)    & 61 & 327 & 127 & 61  \\  
			Disk size (KB)     & 236 & 10535 & 513 &  236 \\
			GPU Memory (GB) & 1.98 & 1.52 & 1.90 & 2.34  \\ \bottomrule
		\end{tabular}
	}
   	% \vspace{-0.6cm}
\end{table}

\textbf{Saga barely increases additional training overhead.}
We can see that the training time and memory consumption of Saga are slightly higher than those of LIMU, which is due to the inclusion of multiple pre-training tasks in Saga. Nonetheless, the training latency of Saga with a batch size of 32 is only 56ms, and the increase in memory consumption compared to LIMU is only 18\%. Such an increase in overhead is clearly acceptable. The parameter count and disk space consumption of Saga are consistent with those of LIMU, as the multiple pre-training tasks designed by Saga do not introduce any additional model structures. These demonstrate the high feasibility of the proposed Saga method.

\subsubsection{Inference Latency}

\begin{figure}[]
	\centering
	\includegraphics[width=0.85\linewidth]{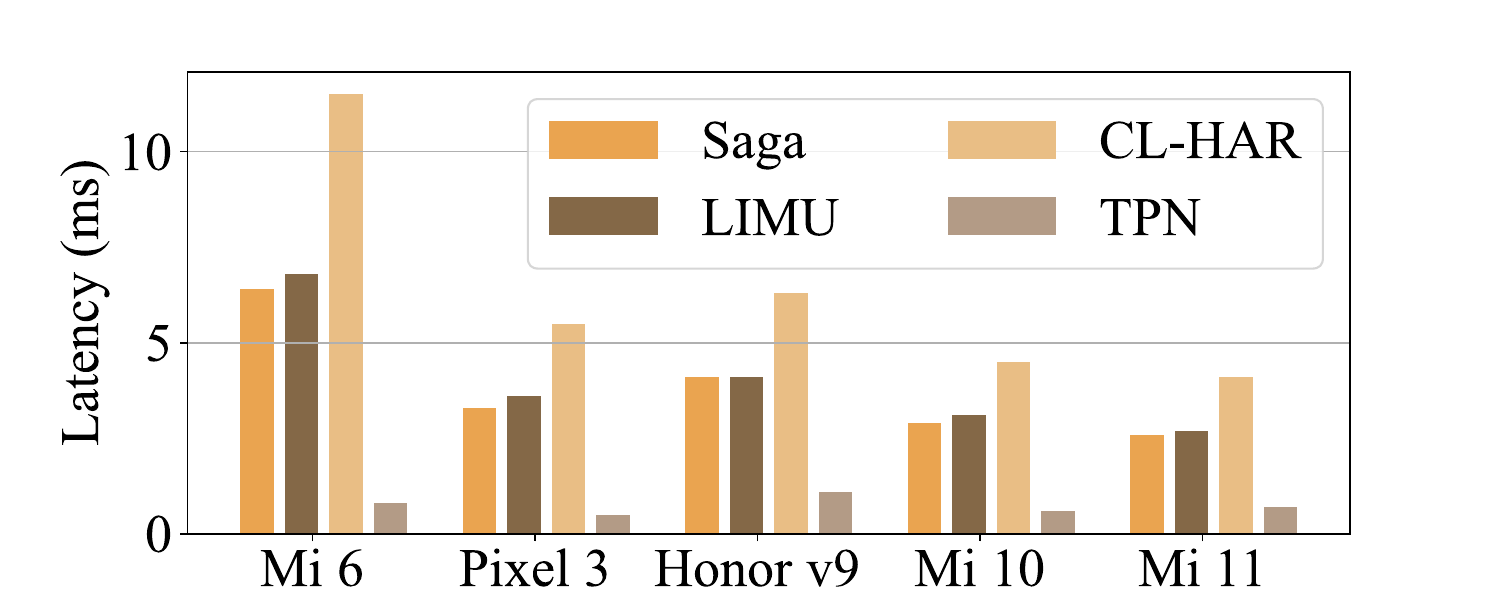}
   % \vspace{-0.4cm}
	\caption{Inference latency of different methods on various mobile phones, where all latencies are less than 12 ms.}
	\label{fig:infer}
  %  \vspace{-0.6cm}
\end{figure}

We further evaluate the inference latency of Saga on various mobile phones. Figure \ref{fig:infer} shows the inference latency of all candidate methods when performing inference on data from two triaxial sensors for a single window length of 120 (\emph{i.e.}, data with a dimension of $1 \times 120 \times 6$). To reduce measurement errors, the latency of each method is measured 10 times and the average values are recorded.

\textbf{The inference latency of Saga is comparable to existing methods.}
We can see that the inference latency of Saga is not higher than that of LIMU, as Saga only adds pre-training tasks without introducing additional computational branches. Although the inference latency of TPN is significantly lower than other methods, it can be seen from Section \ref{sec:perf} that the inference accuracy of TPN is notably lower than that of other candidate methods. In addition, even on the lowest-end devices, Saga achieves a latency of less than 7ms, demonstrating its efficiency during inference and high feasibility for deployment on mobile devices.

\section{Conclusion}

User perception based on IMU data has been widely used for many applications. However, the labelling of IMU data is expensive. As a result, the labelled IMU data is usually very few, resulting in a practical problem, called IMU-based User perception (IUP) problem.
In this paper, Saga is proposed to solve the IUP problem. To this end, four different pre-training tasks targeting four distinct levels of semantics in IMU data are designed to learn a better representation of IMU data. A weight search algorithm based on Bayesian Optimization is designed to efficiently search for the weights of different pre-training tasks. We evaluated the performance of Saga on three tasks from three datasets. Experiments show that the performance of Saga generally surpasses existing state-of-the-art methods. When only using about 100 training samples per class, Saga can achieve a relative accuracy of over 90\% compared with the best results using all labelled IMU data. We believe that Saga can further promote the analysis and understanding of the semantics of IMU data.

\section*{Acknowledgments}
This work was supported in part by the Natural Science Foundation of China (Grants No. 62432008, 62472083), Natural Science Foundation of Shanghai (Grant No. 22ZR1400200) and Ant Group Research Fund (Grant No. 2021110892158).

% \newpage

\bibliographystyle{IEEEtran}
\bibliography{base.bib}
\end{document}